\documentclass[lettersize,journal]{IEEEtran}

\usepackage{graphicx}
\usepackage{amsmath}
\usepackage{amsfonts}
\usepackage{multirow}
\usepackage{cite}
\usepackage[]{hyperref}

\title{
Structured Residual Dynamics Augmented Predictive Control for Indirect Drone-Ball Manipulation
}

\author{Mingjiang Liu$^{1}$, Xiaozhen Zhang$^{1}$, Stephen L. Smith$^{2}$, and Hailong Huang$^{1}$
\thanks{$^{1}$Mingjiang Liu, Xiaozhen Zhang, and Hailong Huang are with the Department of Aeronautical and Aviation Engineering, The Hong Kong Polytechnic University, Hong Kong, China (e-mail: mingjiangaae.liu@connect.polyu.hk; xiaozhen.zhang@polyu.edu.hk; hailong.huang@polyu.edu.hk).}
\thanks{$^{2}$Stephen L. Smith is with the Department of Electrical and Computer Engineering, University of Waterloo, Waterloo, Canada (e-mail: stephen.smith@uwaterloo.ca).}
\thanks{This work has been submitted to the IEEE for possible publication. Copyright may be transferred without notice, after which this version may no longer be accessible.}
}

\begin{document}

% Apply IEEE author-list formatting before the first citation.
\bstctlcite{IEEEreferenceControl}

\maketitle

%==============================================================================
\begin{abstract}
Indirect drone-ball manipulation requires controlling a freely rolling ball through drone-driven beam motion.
This paper presents a nonlinear model predictive control (NMPC) framework for time-varying ball trajectory tracking.
The nominal ball-beam dynamics are augmented by a first-order model of the drone velocity response with command dead time.
A structured ball-acceleration residual is further introduced, motivated by position-dependent and dissipative patterns in the model mismatch.
The resulting dynamics are used in both a state predictor for delay compensation and NMPC.
Using the nominal dynamics, NMPC outperforms a nonlinear feedforward-feedback controller in simulation.
Adding the structured residual to the nominal dynamics improves multi-step prediction on held-out real-world data.
In physical experiments, NMPC using the residual-augmented dynamics reduces overall steady-state error in constant-reference balancing by $50.7\%$ and mean absolute error in dynamic tracking by $37.2\%$ relative to NMPC using the nominal dynamics.
Although the model augmented with a multilayer perceptron (MLP) residual achieves lower offline prediction errors, NMPC using the structured residual achieves better overall closed-loop accuracy and more reliable execution than NMPC using the MLP residual.
% Local sensitivity analysis examines possible explanations for this discrepancy.
\end{abstract}

% Initial Note to Practitioners draft.
\begingroup
\renewcommand{\abstractname}{Note to Practitioners}
\begin{abstract}
Automated handling systems can guide ungrasped objects by moving their supports, offering a potential alternative for positioning and transport when direct gripping is impractical.
This work studies such indirect manipulation using a tethered drone that tilts a beam to guide a freely rolling ball along a desired trajectory.
Accurate control must account for command delays, gradual actuation response, and differences between the mechanical model and actual ball motion.
The proposed controller accounts for the actuator response when predicting future motion and uses a model correction learned before deployment to improve these predictions.
Experiments show better overall accuracy in constant-reference balancing and dynamic tracking, highlighting two key practical lessons.
First, learned residual dynamics can improve control accuracy by compensating for model mismatch. 
Second, the structure of these residuals also matters: incorporating physical priors can benefit closed-loop control 
even when the resulting model does not achieve the lowest offline prediction error.
\end{abstract}
\endgroup


% Task illustration spanning both columns.
\begin{figure*}[t]
    \centering
    \includegraphics[width=0.9\textwidth]{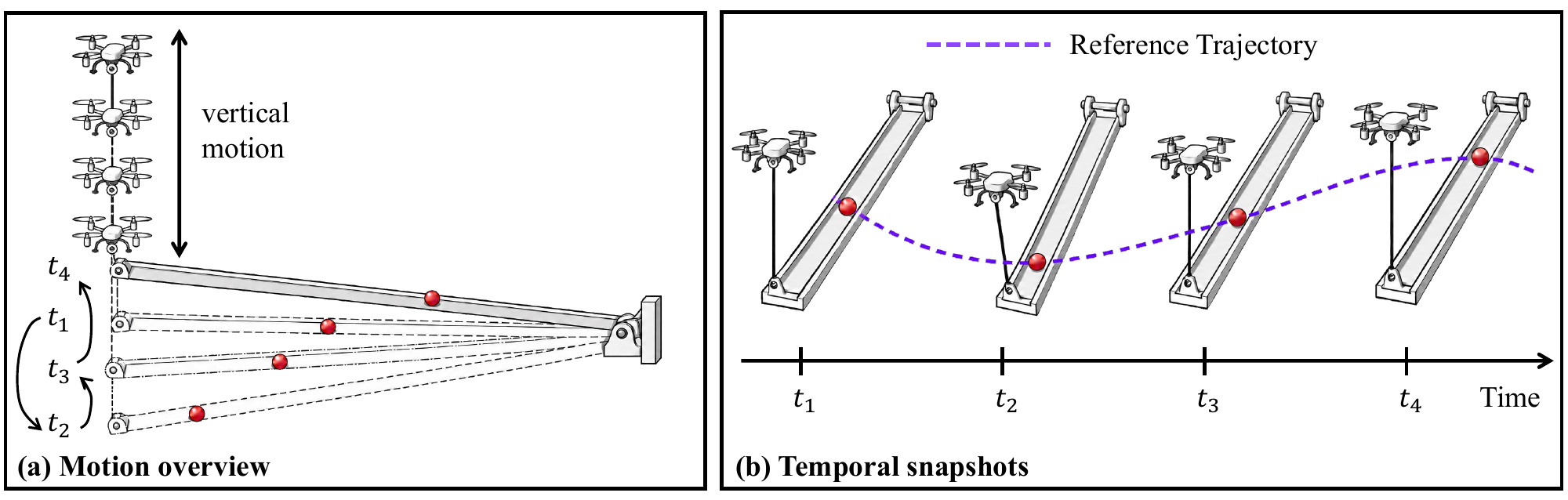}
    \caption{Task illustration of the indirect drone-ball trajectory tracking. Vertical motion of the tethered drone changes the beam
    inclination about a fixed pivot to guide the freely rolling ball. (a) Motion overview with successive
    beam configurations and ball positions overlaid in a shared spatial view. (b) Snapshots of the same system
    at $t_1$--$t_4$, separated horizontally to show temporal evolution. The purple dashed curve
    represents the time-varying ball-position reference.}
    \label{fig:task_illustration}
\end{figure*}

%==============================================================================
\section{INTRODUCTION}
\IEEEPARstart{B}{eyond} mobility and sensing, aerial robots are
increasingly expected to physically interact with and alter their
environments through manipulation. Aerial manipulation is therefore
a key direction for broadening the operational scope and practical
utility of aerial robots~\cite{ollero2022past}.
Recent work has advanced direct aerial manipulation in agility,
versatility, and autonomy. Representative systems have demonstrated
high-speed grasping with a soft gripper and onboard
perception~\cite{2024_npj_robotics_aerial_grasp}, enabled writing and
pick-and-place through a common end-effector-centric
framework~\cite{he2025flyingHand}, and integrated flight and
manipulation for autonomous grasping, perching, door opening, and
transport~\cite{wu2026handLike}. These systems primarily manipulate
objects by holding them with an onboard gripper or by applying
contact forces directly through an end effector.

Tool use in nature illustrates how acting through an intermediate object can extend an animal's
physical capabilities. New Caledonian crows use hooked tools to retrieve food
both in the wild~\cite{hunt1996manufacture} and in
laboratory experiments~\cite{weir2002shaping}.
These examples motivate a complementary direction for aerial robotics: 
extending physical interaction through tools and intermediate mechanisms.
Tool use has recently been demonstrated in aerial robotics, with
FlyingToolbox enabling midair tool exchange~\cite{cao2025proximal}.
Beyond acquiring and exchanging tools, a central control
problem is how to use a given tool to regulate object motion. This becomes
indirect dynamic aerial manipulation when aerial actuation drives an intermediate mechanism
while the object retains its own dynamics. Aerial-Balance-Bench provides a controlled
and reproducible testbed for indirect dynamic aerial manipulation with a given tool~\cite{liu2026aerialBalanceBench}: a vertically
constrained tethered drone guides the motion of a freely rolling ball by tilting its supporting beam. Although simplified relative to unconstrained flight,
the setup retains the coupling among aerial actuation, tool motion, and object dynamics.

Aerial-Balance-Bench leaves two control questions open~\cite{liu2026aerialBalanceBench}.
First, its reference controllers were evaluated for
constant-reference balancing, which requires bringing the ball
to a specified position and maintaining it there, leaving
dynamic tracking of time-varying ball trajectories unexplored.
Second, although its state predictor compensated for command latency,
control remained based on nominal ball and beam dynamics.
The lower performance reported at off-center targets suggests
limitations in this nominal description.
More generally, model mismatch can lead to persistent tracking
offsets~\cite{morari2012offsetfree} and degrade tracking performance
in predictive control~\cite{torrente2021data}.
Extending Aerial-Balance-Bench to dynamic tracking calls for
accurate prediction of both the actuation response and the
ball's rolling dynamics.

To address these challenges, we first augment the nominal ball-beam dynamics with
an identified first-order model of the drone's vertical velocity response
with command dead time (FOPDT).
However, this augmentation addresses only the low-level actuation response; 
a mismatch remains between the ball motion predicted by the nominal model and the ball’s actual motion in the physical system.
We then introduce a structured ball-acceleration residual with a
position-dependent beam-profile term and a friction-inspired dissipative
term, and fit its parameters to real-system data.
These residual-augmented dynamics are used in two key components of the predictive
control framework: a state predictor for command-delay compensation and
a nonlinear model predictive controller (NMPC).
The state predictor advances the measured state through pending commands over
the delay interval.
NMPC then optimizes future command increments from the delay-compensated
state using time-aligned reference preview.

The main contributions are summarized as follows:
\begin{enumerate}
    \item We develop a structured ball-acceleration residual
    comprising a position-dependent beam-profile correction and
    a friction-inspired dissipative term to address model mismatch
    in nominal ball dynamics.

    \item Building on augmented nominal dynamics with an identified
    low-level actuation response, we incorporate the structured
    residual consistently into both a delay-compensating state
    predictor and NMPC. The resulting framework handles
    constant-reference balancing and dynamic tracking within a
    unified formulation.

    \item We evaluate the predictive and closed-loop benefits of
    the structured residual through component ablations, comparisons
    with generic residual models, and physical control experiments.
    The study demonstrates improvements over nominal prediction and
    control and shows that lower offline prediction error need not
    imply better closed-loop accuracy.
\end{enumerate}

%==============================================================================
\section{Related Work}
\label{sec:related_work}

\subsection{Ball-and-Beam Control with Robotic Actuation}
The ball-and-beam system has long served as a benchmark for nonlinear and
underactuated control, with early experimental platforms dating back to the
1970s~\cite{wellstead1978ballbeam}. In conventional configurations, a dedicated
actuator directly regulates beam motion, while the ball is controlled indirectly
through its rolling dynamics. This compact actuation path has supported the
development of nonlinear~\cite{hauser1992ballbeam}, passivity-based~\cite{serra2019nonprehensile}, 
and adaptive control methods~\cite{li2025adaptive}, making the
system a useful idealized testbed but abstracting away the additional actuation
layers present in robotic platforms. Practical extensions have considered
nonidealities such as model uncertainty~\cite{nezhad2025ballbeam} and communication latency~\cite{lyu2025latency}, 
as well as alternative beam actuators such as magnetic
suspension~\cite{ker2007magnetic}. Nevertheless, beam motion in these systems is still driven
by dedicated actuators.

Robotic extensions use general-purpose manipulators to move the beam,
including balance control with a redundant manipulator~\cite{ryu2011redundant},
sensorimotor reinforcement learning for ball-position
control~\cite{ghadirzadeh2016sensorimotor}, and a learning-based hierarchical
control framework~\cite{jin2023hierarchical}.
In these settings, beam motion is realized through robot kinematics and
low-level control.
These configurations bring the ball-and-beam task closer to robotic manipulation, 
but remain based on ground manipulators.
Aerial-Balance-Bench~\cite{liu2026aerialBalanceBench} extends this
progression to aerial actuation, where drone motion is transmitted through the
rope-beam geometry before affecting the rolling ball. In its velocity-command
realization, the command is additionally mediated by the drone's low-level flight
response. Compared with conventional and manipulator-driven systems, this longer
and dynamic actuation path introduces additional modeling and control challenges.

\subsection{Model Mismatch Compensation in Robotic Control}
Model mismatch is common in robotic systems because effects such as contact,
friction, aerodynamics, and actuator dynamics are difficult to capture completely
with analytical models. One approach is to compensate for such mismatch at the
control-policy level. Residual policies can augment conventional feedback controllers to handle
contact and friction effects in robotic assembly~\cite{johannink2019residual}
and compensate for aerodynamic disturbances during close-proximity
quadcopter flight through thrust and body-rate
corrections~\cite{zhang2025proxfly}.
Related approaches learn residual joint-position targets to address
reference-robot dynamics mismatch in long-horizon humanoid motion
tracking~\cite{sun2025robotdancing}, or blend learned corrections with
MPC outputs at the torque level to improve robustness in legged
locomotion~\cite{jeon2026residualmpc}.
These approaches preserve useful prior behavior without explicitly modeling the
missing dynamics. However, learning an effective residual policy still requires
experience that exposes the relevant mismatch, either from physical interaction
or from simulation that sufficiently reproduces it, which may be costly or
difficult to obtain. 

An alternative is to compensate for mismatch at the model level.
Offset-free MPC addresses steady-state tracking offsets caused by persistent
model mismatch through augmented prediction models and online estimation,
using disturbance-state augmentation~\cite{morari2012offsetfree} or delta-input
formulations combined with state estimation~\cite{yigit2023macromini}.
Adaptive~\cite{o2022neural} or learning-based
approaches~\cite{zhou2025simultaneous} can further update unknown dynamics during
operation to accommodate changing conditions. Such online methods improve adaptability, 
but introduce additional estimation or model updating into the control loop and 
may require sufficiently informative data.
When the dominant mismatch is sufficiently repeatable, residual dynamics can
instead be learned offline and kept fixed during deployment, as demonstrated for
aerodynamic interactions~\cite{shi2021neural} and residual models embedded in
predictive control~\cite{torrente2021data}.
These approaches retain nominal physics while learning the unmodeled component
through Gaussian process or neural representations. Beyond retaining the nominal model,
a further design choice is how to encode physical structure within the learned
correction itself. Related work on physics-informed dynamics
learning~\cite{lutter2023physics} and port-Hamiltonian neural
models~\cite{duong2024porthamiltonian} suggests that incorporating physical
structure can improve physical consistency and suitability for control,
motivating the use of physical priors in residual modeling.

%==============================================================================
\section{Problem Formulation}
\label{sec:problem_formulation}
We consider the drone-ball system illustrated in
Fig.~\ref{fig:task_illustration}, in which a rope connects a
vertically constrained drone to one end of a beam while the
opposite end remains fixed.
Vertical drone motion changes the beam inclination and thereby
guides a freely rolling ball along the beam.
Following Aerial-Balance-Bench~\cite{liu2026aerialBalanceBench},
this configuration focuses on the coupling among drone actuation,
beam inclination, and ball motion.
The scope is restricted to vertical drone motion and ball motion
along the beam rather than general three-dimensional motion
of the drone and ball.

\begin{figure}[t]
    \centering
    \includegraphics[width=0.9\columnwidth]{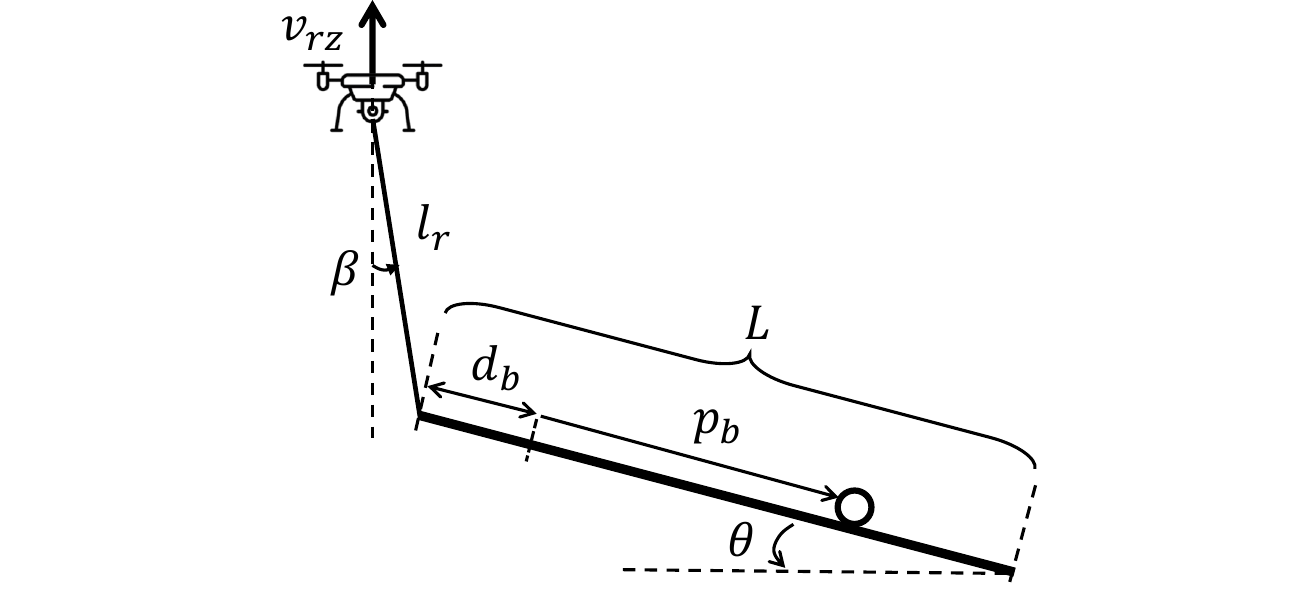}
    \caption{Geometry and coordinate conventions of the drone-ball
    system. }
    \label{fig:system_geometry}
\end{figure}

Fig.~\ref{fig:system_geometry} defines the geometry and coordinate
conventions used in the model. Here, $L$ and $l_r$ denote the beam and rope lengths,
while $\theta$ and $\beta$ denote the beam inclination from
horizontal and the rope angle from vertical, respectively.
The ball position $p_b$ increases toward the fixed end from an
origin offset by $d_b$ from the rope attachment point along the beam.
The corresponding velocities are $v_b=\dot p_b$ and
$\omega=\dot\theta$, and the drone vertical velocity $v_{rz}$
is positive upward.

We use a hierarchical velocity-control interface: the high-level
controller updates the commanded drone vertical velocity $c_z$,
and the low-level controller tracks this command, producing the
actual velocity $v_{rz}$. The high-level controller uses
measurements or estimates of the physical state
\begin{equation}
    \boldsymbol{\xi}(t)
    =[p_b(t),v_b(t),\theta(t),v_{rz}(t)]^\top.
    \label{eq:physical_state_definition}
\end{equation}
At sampling time $t_k=kh$, where $h$ is the sampling interval,
the controller outputs a velocity command increment:
\begin{equation}
\begin{aligned}
    u_k &= \Delta c_{z,k},\\
    c_z(t_k^+) &= c_z(t_k^-)+u_k.
\end{aligned}
    \label{eq:control_interface_update}
\end{equation}
Here, $t_k^-$ and $t_k^+$ denote the
instants immediately before and after the update. The updated command is
held until the next sampling time. The commanded and actual velocities
differ because of the low-level actuation response.

The task is to control the ball position $p_b(t)$ to follow a
prescribed reference $p_g(t)$, expressed in the same beam coordinate.
A constant reference defines constant-reference balancing,
whereas a time-varying reference defines dynamic tracking.
Both tasks are therefore expressed within the same tracking formulation. 
The reference is  assumed to be available at the current time and over 
the preview interval required by the controller.
Fig.~\ref{fig:task_illustration} illustrates dynamic
tracking through successive beam configurations and ball positions.

%==============================================================================
\section{Structured Residual-Aware Predictive Control}
\label{sec:residual_aware_control}
\subsection{Augmented Nominal Dynamics}
\label{sec:augmented_nominal_dynamics}
The nominal prediction model augments the mechanical dynamics
of the drone-ball system with the low-level actuation response.
From the geometry shown in Fig.~\ref{fig:system_geometry}, the rope angle
and beam angular velocity are given by
\begin{equation}
\begin{aligned}
    \beta(\theta)&=\arcsin\!\left[\frac{L}{l_r}(1-\cos\theta)\right],\\
    \omega(\theta,v_{rz})&=-\frac{v_{rz}\cos\beta(\theta)}
    {L\cos\!\left(\beta(\theta)-\theta\right)}.
\end{aligned}
    \label{eq:beam_geometry}
\end{equation}
Let $m_b$, $r_b$, and $J_b$ denote the ball mass, radius,
and moment of inertia about its rolling axis through the
center, respectively. With $\gamma=m_b/(m_b+J_b/r_b^2)$,
the ball acceleration predicted by the nominal mechanical model is
\begin{equation}
    a_b^{\mathrm{nom}}(p_b,\theta,v_{rz})
    =\gamma(p_b+d_b-L)\omega^2(\theta,v_{rz})-\gamma g\sin\theta,
    \label{eq:nominal_ball_acceleration}
\end{equation}
where $g$ is the gravitational acceleration.

The mechanical model depends on the actual drone velocity $v_{rz}$,
whereas the control input updates the commanded velocity $c_z$.
To relate these two quantities, we model the low-level actuation response
as a FOPDT:
\begin{equation}
\begin{aligned}
    c_z^d(t)&=c_z(t-T_d),\\
    \tau\dot v_{rz}(t)&=Kc_z^d(t)+b-v_{rz}(t).
\end{aligned}
    \label{eq:fopdt_actuation_response}
\end{equation}
Here, $T_d$ is the command dead time, $\tau$ describes the gradual velocity
response after a command takes effect, and $K$ and $b$ are the steady-state
gain and velocity bias, respectively.

Combining the mechanical dynamics and actuation response gives the
augmented state
\begin{equation}
    \mathbf x(t)
    =\begin{bmatrix}
        p_b(t)&v_b(t)&\theta(t)&v_{rz}(t)&c_z(t)
    \end{bmatrix}^{\!\top}.
    \label{eq:augmented_nominal_state}
\end{equation}
Between command updates, the augmented nominal dynamics are
\begin{equation}
    \dot{\mathbf x}(t)
    =\mathbf f^{\mathrm{nom}}
    \!\left(\mathbf x(t),c_z^d(t)\right)
    =
    \begin{bmatrix}
        v_b(t)\\[1mm]
        a_b^{\mathrm{nom}}\!\left(p_b(t),\theta(t),v_{rz}(t)\right)\\[1mm]
        \omega\!\left(\theta(t),v_{rz}(t)\right)\\[1mm]
        \displaystyle\frac{Kc_z^d(t)+b-v_{rz}(t)}{\tau}\\[2mm]
        0
    \end{bmatrix}.
    \label{eq:augmented_nominal_dynamics}
\end{equation}
The second and third rows describe the coupled ball and beam motion, and
the fourth represents the first-order velocity response. The last row
holds the command constant between sampling instants; its discrete update
is given by \eqref{eq:control_interface_update}. The delayed command $c_z^d(t)$ 
is retrieved from the stored command history.

For discrete-time prediction, the effective delayed command
$c_{z,k}^d$ is held constant over each interval of duration $h$.
With $\bar v_{rz,k}=Kc_{z,k}^d+b$, the first-order velocity
response is
\begin{equation}
    v_{rz,k}(s)=\bar v_{rz,k}
    +(v_{rz,k}-\bar v_{rz,k})e^{-s/\tau},
    \quad s\in[0,h],
    \label{eq:exact_fopdt_response}
\end{equation}
where $v_{rz,k}$ is the initial velocity.
We set $v_{rz,k+1}=v_{rz,k}(h)$ and integrate
$p_b$, $v_b$, and $\theta$ using fourth-order Runge--Kutta
(RK4), evaluating this response at $s=0$, $h/2$, and $h$
for the corresponding stages.

\subsection{Structured Residual Dynamics}\label{structured_residual_model}
The augmented nominal model captures the mechanical dynamics and low-level
actuation response, but does not fully describe the position-dependent and
dissipative effects in ball motion. We account for these effects through a
structured residual correction to ball acceleration, while retaining the
nominal geometry and identified actuation response:
\begin{equation}
    a_b^{\mathrm{res}}(p_b,v_b,\theta,v_{rz})=a_b^{\mathrm{nom}}(p_b,\theta,v_{rz})+r_a(p_b,v_b,\theta).
    \label{eq:structured_residual_dynamics}
\end{equation}
Here, $a_b^{\mathrm{res}}$ is the corrected ball acceleration and $r_a$
is the learned acceleration residual. The residual is structured as the sum of a beam-profile correction
and a friction-inspired dissipative term, with a smooth bound
applied to the resulting acceleration correction:
\begin{equation}
\begin{aligned}
    r_a^{\mathrm{raw}}&=r_{\mathrm{prof}}+r_{\mathrm{diss}},\\
    r_a&=\bar r_a\tanh\!\left(\frac{r_a^{\mathrm{raw}}}{\bar r_a}\right),
\end{aligned}
    \label{eq:structured_residual_components}
\end{equation}
where $r_a^{\mathrm{raw}}$ is the sum before output bounding and $\bar r_a>0$
sets the residual-acceleration limit. The outer hyperbolic tangent keeps
the applied residual bounded and smooth.

The beam-profile component represents a position-dependent correction
$\delta\theta(p_b)$ to the effective local slope:
\begin{equation}
    r_{\mathrm{prof}}(p_b,\theta)
    =-\gamma g\!\left[
    \sin\!\bigl(\theta+\delta\theta(p_b)\bigr)-\sin\theta\right].
    \label{eq:profile_residual}
\end{equation}
This term modifies the gravity projection through an effective slope change.
The dissipative component is
\begin{equation}
    r_{\mathrm{diss}}(p_b,v_b)
    =-f_c(p_b)\tanh\!\left(\frac{v_b}{v_0}\right)-f_v(p_b)v_b.
    \label{eq:dissipative_residual}
\end{equation}
Here, $f_c(p_b)$ is a Coulomb-like acceleration magnitude, $f_v(p_b)$ is a
viscous coefficient, and $v_0>0$ sets the smooth transition around zero
velocity. Nonnegative $f_c$ and $f_v$ ensure that the dissipative component
opposes ball motion.

We parameterize $\delta\theta$, $f_c$, and $f_v$ using shared normalized
Gaussian radial basis functions (RBFs). Let $M$ denote the number of bases,
$\mu_i$ their centers, $\sigma_p$ their common width, and $\epsilon>0$ a
small normalization constant. Then
\begin{equation}
\begin{aligned}
    \phi_i(p_b)
    &=\exp\!\left[-\frac{(p_b-\mu_i)^2}{2\sigma_p^2}\right],\\
    \bar\phi_i(p_b)
    &=\frac{\phi_i(p_b)}
    {\epsilon+\sum_{j=1}^{M}\phi_j(p_b)},\\
    \delta\theta(p_b)
    &=\sum_{i=1}^{M}w_i^p\bar\phi_i(p_b),\\
    \begin{bmatrix}f_c(p_b)\\ f_v(p_b)\end{bmatrix}
    &=\sum_{i=1}^{M}\bar\phi_i(p_b)
    \begin{bmatrix}w_i^c\\ w_i^v\end{bmatrix}.
\end{aligned}
    \label{eq:normalized_position_rbf}
\end{equation}
Here, $\phi_i$ is the $i$th Gaussian basis and $\bar\phi_i$ its normalized
value, for $i=1,\ldots,M$. The centers are uniformly distributed over the
valid beam region.
The coefficients $w_i^p$, $w_i^c$, and $w_i^v$ parameterize the slope
correction, Coulomb-like magnitude, and viscous coefficient, respectively,
subject to
\begin{equation}
    |w_i^p|\leq\bar\delta_\theta,\qquad
    0\leq w_i^c\leq\bar f_c,\qquad
    0\leq w_i^v\leq\bar f_v.
    \label{eq:structured_coefficient_bounds}
\end{equation}
Here, $\bar\delta_\theta$, $\bar f_c$, and $\bar f_v$ bound the slope
correction and the two dissipative coefficients, respectively.
These components are structured approximations, not identified measurements
of the beam profile or friction forces.

The residual parameters are learned offline from real closed-loop data.
We estimate ball acceleration $\hat a_b$ from the second time derivative
of the smoothed ball-position signal and subtract the nominal acceleration
to obtain the training target:
\begin{equation}
    r_a^{\mathrm{data}}(t)
    =\hat a_b(t)-a_b^{\mathrm{nom}}
    \bigl(p_b(t),\theta(t),v_{rz}(t)\bigr).
    \label{eq:residual_acceleration_target}
\end{equation}
The training objective combines pointwise fit, integral consistency, spatial
smoothness, and causal rollout:
\begin{equation}
    \mathcal L=\lambda_{\mathrm{pt}}\mathcal L_{\mathrm{pt}}
    +\mathcal L_{\mathrm{int}}
    +\lambda_{\mathrm{sm}}\mathcal L_{\mathrm{sm}}
    +\lambda_{\mathrm{roll}}\mathcal L_{\mathrm{roll}}.
    \label{eq:structured_residual_training_loss}
\end{equation}
The nonnegative weights $\lambda_{\mathrm{pt}}$, $\lambda_{\mathrm{sm}}$,
and $\lambda_{\mathrm{roll}}$ set the relative contributions of pointwise
fit, coefficient smoothness, and rollout accuracy, respectively. We use the smooth L1 penalty
\begin{equation}
    \rho_\delta(e)=
    \begin{cases}
        e^2/(2\delta), & |e|<\delta,\\
        |e|-\delta/2, & |e|\geq\delta,
    \end{cases}
    \label{eq:smooth_l1_penalty}
\end{equation}
where $e$ is a scalar error and $\delta>0$ is the transition threshold.
We use $\delta_a$ for both pointwise acceleration and integral-consistency
errors, and $\delta_r$ for normalized rollout errors. Let
$\langle\cdot\rangle_{\mathcal D}$ and $\langle\cdot\rangle_{\mathcal W}$
denote averages over sampled training points and windows, respectively.
The pointwise term fits the bounded residual to its acceleration target:
\begin{equation}
    \mathcal L_{\mathrm{pt}}
    =\left\langle\rho_{\delta_a}
    \bigl(r_a-r_a^{\mathrm{data}}\bigr)\right\rangle_{\mathcal D}.
    \label{eq:pointwise_residual_loss}
\end{equation}

The integral term compares the average predicted acceleration with the
observed velocity change over a data window. For local window indices
$j=0,\ldots,H$, let $\Delta t_j$ denote the sampling intervals and
$\tilde v_{b,j}$ the ball velocities estimated from recorded positions
using the first derivative of a Savitzky-Golay local polynomial fit. Define
\begin{equation}
\begin{aligned}
    T_H&=\sum_{j=0}^{H-1}\Delta t_j,\\
    A_H&=\sum_{j=0}^{H-1}
    \frac{a_j^{\mathrm{res}}+a_{j+1}^{\mathrm{res}}}{2}\Delta t_j,\\
    E_H^{\mathrm{int}}
    &=\frac{A_H-(\tilde v_{b,H}-\tilde v_{b,0})}{T_H},\\
    \mathcal L_{\mathrm{int}}
    &=\sum_{H\in\mathcal H_{\mathrm{int}}}\alpha_H
    \left\langle\rho_{\delta_a}(E_H^{\mathrm{int}})
    \right\rangle_{\mathcal W}.
\end{aligned}
    \label{eq:integral_consistency_loss}
\end{equation}
The trapezoidal sum $A_H$ approximates
the acceleration integral, $T_H$ is the window duration, and
$E_H^{\mathrm{int}}$ is the average acceleration mismatch inferred from
that integral and the observed velocity change. The set
$\mathcal H_{\mathrm{int}}$ contains the
window lengths, measured in sampling intervals, and $\alpha_H$ is the weight for length $H$.

The smoothness term penalizes first and second differences of normalized
RBF coefficients. Let $\mathbf w^q=[w_1^q,\ldots,w_M^q]^\top$, with
$q\in\{p,c,v\}$ indexing the profile, Coulomb-like, and viscous coefficients.
Define $\tilde{\mathbf w}^p=\mathbf w^p/\bar\delta_\theta$,
$\tilde{\mathbf w}^c=\mathbf w^c/\bar f_c$, and
$\tilde{\mathbf w}^v=\mathbf w^v/\bar f_v$. With
$\Delta w_i=w_{i+1}-w_i$ and
$\Delta^2w_i=w_{i+2}-2w_{i+1}+w_i$,
\begin{equation}
    \mathcal L_{\mathrm{sm}}=
    \sum_{q\in\{p,c,v\}}\left[
    \eta_1\frac{\|\Delta\tilde{\mathbf w}^q\|_2^2}{M-1}
    +\eta_2\frac{\|\Delta^2\tilde{\mathbf w}^q\|_2^2}{M-2}
    \right].
    \label{eq:rbf_smoothness_loss}
\end{equation}
Here, $\eta_1$ and $\eta_2$ are nonnegative weights for the first- and
second-difference penalties, respectively; the denominators average over
the available differences. This regularizer discourages abrupt spatial
variation of the learned corrections.

For the rollout term, let $\mathbf y=[p_b,v_b]^\top$ be a target state and
$\hat{\mathbf y}=[\hat p_b,\hat v_b]^\top$ its model prediction. Define
\begin{equation}
    \ell(\hat{\mathbf y},\mathbf y)
    =\rho_{\delta_r}\!\left(\frac{\hat p_b-p_b}{s_p}\right)
    +\rho_{\delta_r}\!\left(\frac{\hat v_b-v_b}{s_v}\right),
    \label{eq:normalized_rollout_error}
\end{equation}
where $s_p>0$ and $s_v>0$ are position and velocity error scales,
respectively, making both penalty arguments dimensionless.
Starting from a causal state estimate and its pending-command history, we
propagate the model using recorded commands, recomputing residual features
from predicted states at each integration stage. Let $\hat{\mathbf y}_H$
be the resulting prediction, $\tilde{\mathbf y}_H$ the smoothed data target,
and $\mathbf y_H^c$ the state estimate obtained using only measurements
available up to that time. The loss is
\begin{equation}
    \mathcal L_{\mathrm{roll}}=\sum_{H\in\mathcal H_{\mathrm{roll}}}\beta_H
    \big\langle\ell(\hat{\mathbf y}_H,\tilde{\mathbf y}_H)
    +\eta_c\ell(\hat{\mathbf y}_H,\mathbf y_H^c)\big\rangle_{\mathcal W}.
    \label{eq:causal_rollout_training_loss}
\end{equation}
The targets are used only to evaluate the loss; the rollout proceeds
without replacing predicted states with measured or smoothed states.
The set $\mathcal H_{\mathrm{roll}}$ contains the prediction horizons in
sampling steps. The nonnegative weight $\beta_H$ selects the contribution
of horizon $H$, while $\eta_c$ weights the causal-estimate target relative
to the smoothed target.

Training proceeds in two stages: the first disables the rollout term,
and the second enables it. After each parameter update, the coefficients
are projected onto the feasible set defined by
\eqref{eq:structured_coefficient_bounds}.

\subsection{Residual-Aware Predictive Control}
\begin{figure}[t]
    \centering
    \includegraphics[width=0.95\columnwidth]{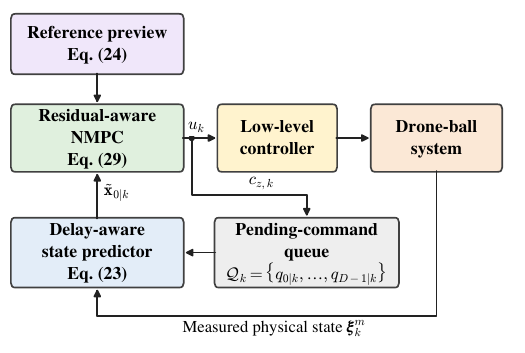}
    \caption{Residual-aware realization of the delay-aware control structure. The predictor advances the
    measured state through the ordered pending-command queue over the pure-delay interval, and NMPC starts from
    the resulting delay-compensated state with a time-aligned reference preview. Both the predictor and NMPC use
    the same structured residual dynamics.}
    \label{fig:residual_aware_control_structure}
\end{figure}

The predictive control framework integrates a state predictor for
command-delay compensation with a high-level NMPC controller,
as illustrated in Fig.~\ref{fig:residual_aware_control_structure}.
The predictor propagates the measured state through the pending
commands, and NMPC optimizes future commands from the resulting
delay-compensated state.
Both components use the same augmented dynamics with the structured
residual, ensuring consistent models across the two prediction stages.
Command dead time is handled by the predictor and is not included
again in the NMPC horizon.

\textit{1) Delay-aware state prediction:}
The predictor propagates the physical state defined in
\eqref{eq:physical_state_definition}. Let $\boldsymbol{\xi}_k^m$ denote
the state obtained from measurements at time $t_k$;
the superscript $m$ distinguishes this supplied state from a model
prediction. We approximate the command dead time by $D$ sampling intervals,
so that $T_d\approx Dh$. The ordered queue
$\mathcal Q_k=\{q_{0|k},\ldots,q_{D-1|k}\}$ contains the pending velocity
commands, from oldest to newest. Each $q_{j|k}$ is known at time $t_k$ and
serves as the held effective input $c_z^d$ over
$[t_{k+j},t_{k+j+1})$; it is a velocity command rather than an increment
$u_k$. Let $\hat{\boldsymbol{\xi}}_{k+j|k}$ denote the state predicted for
time $t_{k+j}$ using information available at $t_k$. The predictor computes
\begin{equation}
\begin{aligned}
    \hat{\boldsymbol{\xi}}_{k|k}
    &=\boldsymbol{\xi}_k^m,\\
    \hat{\boldsymbol{\xi}}_{k+j+1|k}
    &=\mathbf G_h^{\mathrm{res}}
    \bigl(\hat{\boldsymbol{\xi}}_{k+j|k},q_{j|k}\bigr),
    \quad j=0{:}D-1,\\
    \tilde{\mathbf x}_{0|k}
    &=\begin{bmatrix}
        \hat{\boldsymbol{\xi}}_{k+D|k}^{\top}&c_z(t_k^-)
    \end{bmatrix}^{\!\top}.
\end{aligned}
    \label{eq:delay_state_prediction}
\end{equation}
The map $\mathbf G_h^{\mathrm{res}}$ advances the four-dimensional physical
state by one sampling interval under a held effective velocity command.
It integrates the first four rows of \eqref{eq:augmented_nominal_dynamics},
with $q_{j|k}$ substituted for $c_z^d$ and $a_b^{\mathrm{nom}}$ replaced by
$a_b^{\mathrm{res}}$ from \eqref{eq:structured_residual_dynamics}.
The integration uses the analytical velocity response in
\eqref{eq:exact_fopdt_response} and evaluates $r_a$ at every RK4 stage.
The vector $\tilde{\mathbf x}_{0|k}$ initializes NMPC: its physical
component represents the predicted state at $t_k+Dh$, when the updated
command is expected to take effect. Its command component $c_z(t_k^-)$
is the current accumulator that provides the baseline for the new increment
$u_k$, rather than a predicted future command. The model parameters remain
fixed during online prediction.

Let $N$ denote the NMPC prediction horizon in sampling intervals and $i$
the stage index counted from the end of the delay interval. The position
and velocity references for stage $i$, constructed at time $t_k$, are
aligned with the same future time as the predicted physical state:
\begin{equation}
\begin{aligned}
    p_{g,i|k}&=p_g\!\left(t_k+(D+i)h\right),\\
    v_{g,i|k}&=v_g\!\left(t_k+(D+i)h\right),
\end{aligned}
    \label{eq:delay_aligned_reference}
\end{equation}
where $i=0,\ldots,N$. The reference generator supplies the velocity
reference $v_g$, equal to $\dot p_g$ wherever the reference is differentiable;
at transitions in piecewise trajectories, the generator's prescribed
stage value is used. For a constant reference position $p_g^{\mathrm{const}}$, the references
are $p_{g,i|k}=p_g^{\mathrm{const}}$ and $v_{g,i|k}=0$.
The reference preview must therefore extend to $t_k+(D+N)h$.

\textit{2) Residual-aware NMPC:}
Let $\mathbf x_{i|k}$ denote the augmented state at stage $i$ of the
optimization performed at $t_k$, with physical time $t_k+(D+i)h$, and let
$u_{i|k}$ be its commanded velocity increment. Using
\eqref{eq:augmented_nominal_state} and \eqref{eq:control_interface_update},
the prediction model is
\begin{equation}
\begin{aligned}
    \mathbf x_{0|k}&=\tilde{\mathbf x}_{0|k},\\
    \mathbf x_{i+1|k}
    &=\mathbf F_h^{\mathrm{res}}
    \bigl(\mathbf x_{i|k},u_{i|k}\bigr),
    \quad i=0{:}N-1.
\end{aligned}
    \label{eq:residual_aware_prediction_model}
\end{equation}
For $\mathbf x=[\boldsymbol{\xi}^{\top},c_z]^{\top}$, the transition map is
\begin{equation}
    \mathbf F_h^{\mathrm{res}}(\mathbf x,u)
    =\begin{bmatrix}
        \mathbf G_h^{\mathrm{res}}(\boldsymbol{\xi},c_z+u)\\
        c_z+u
    \end{bmatrix}.
    \label{eq:augmented_discrete_transition}
\end{equation}
Thus, $c_{z,i|k}$ is the command accumulator before the stage update,
and $c_{z,i|k}+u_{i|k}$ drives the physical-state transition and becomes
the next command state. The map $\mathbf G_h^{\mathrm{res}}$ uses the
integration scheme defined above, without applying command dead time again.

The stage and terminal cost residuals are
\begin{equation}
    \mathbf e_{i|k}
    =\begin{bmatrix}
        p_{b,i|k}-p_{g,i|k}\\
        v_{b,i|k}-v_{g,i|k}\\
        \theta_{i|k}\\
        \omega_{i|k}\\
        v_{rz,i|k}\\
        c_{z,i|k}+u_{i|k}\\
        u_{i|k}
    \end{bmatrix},\qquad
    \mathbf e_{N|k}^{\mathrm{term}}
    =\begin{bmatrix}
        p_{b,N|k}-p_{g,N|k}\\
        v_{b,N|k}-v_{g,N|k}\\
        \theta_{N|k}\\
        \omega_{N|k}\\
        v_{rz,N|k}\\
        c_{z,N|k}
    \end{bmatrix}.
    \label{eq:nmpc_tracking_residuals}
\end{equation}
Here, the superscript $\mathrm{term}$ denotes the terminal stage, and
$\omega_{i|k}=\omega(\theta_{i|k},v_{rz,i|k})$ is computed from
\eqref{eq:beam_geometry}, rather than treated as an independent state.
The first two entries penalize tracking errors; the remaining entries
penalize beam and drone motion, commanded velocity, and command increments.
Let $\mathbf s_{i|k}^l,\mathbf s_{i|k}^u$ be three-dimensional nonnegative
lower- and upper-bound slack vectors, with components corresponding to
ball position, beam angle, and tracking error. The objective is defined as
\begin{equation}
\begin{aligned}
    J_k
    &=\frac{1}{2}\sum_{i=0}^{N-1}
    \mathbf e_{i|k}^{\top}\mathbf W\mathbf e_{i|k}+\frac{1}{2}(\mathbf e_{N|k}^{\mathrm{term}})^{\top}
    \mathbf W_N\mathbf e_{N|k}^{\mathrm{term}}+J_{\mathrm{slack}},\\
    J_{\mathrm{slack}}
    &=\sum_{i=0}^{N}\Bigl[
    \rho_1\mathbf 1_3^\top(\mathbf s_{i|k}^l+\mathbf s_{i|k}^u)+\frac{\rho_2}{2}\left(\lVert\mathbf s_{i|k}^l\rVert_2^2
    +\lVert\mathbf s_{i|k}^u\rVert_2^2\right)\Bigr],
\end{aligned}
    \label{eq:nmpc_tracking_objective}
\end{equation}
where $\mathbf W\in\mathbb{R}^{7\times7}$ and
$\mathbf W_N\in\mathbb{R}^{6\times6}$ are diagonal weighting matrices
with nonnegative diagonal entries. The nonnegative coefficients $\rho_1$ and $\rho_2$ weight linear
and quadratic slack penalties, respectively; $\mathbf 1_d$ denotes a
$d$-dimensional vector of ones.

The softened quantities are
$\mathbf h_{i|k}=[p_{b,i|k},\theta_{i|k},p_{b,i|k}-p_{g,i|k}]^\top$.
Their lower and upper bound vectors are $\underline{\mathbf h}$ and
$\overline{\mathbf h}$, in the same component order. These bounds permit penalized violations and do not provide
a hard guarantee that the ball remains within the prescribed
position limits.
Define the decision sequences
$\mathbf X_k=\{\mathbf x_{i|k}\}_{i=0}^{N}$,
$\mathbf U_k=\{u_{i|k}\}_{i=0}^{N-1}$, and
$\mathbf S_k^{l,u}=\{\mathbf s_{i|k}^{l,u}\}_{i=0}^{N}$.
The optimal control problem (OCP) is
\begin{equation}
\begin{aligned}
    \min_{\mathbf X_k,\mathbf U_k,\mathbf S_k^l,\mathbf S_k^u}\quad
    &J_k\\
    \mathrm{s.t.}\quad
    &\mathbf x_{0|k}=\tilde{\mathbf x}_{0|k},\\
    &\mathbf x_{i+1|k}=\mathbf F_h^{\mathrm{res}}
    (\mathbf x_{i|k},u_{i|k}),\\
    &-a_{\max}h\leq u_{i|k}\leq a_{\max}h,\\
    &-v_{\max}\mathbf 1_2\leq
    \begin{bmatrix}v_{rz,i|k}\\c_{z,i|k}\end{bmatrix}
    \leq v_{\max}\mathbf 1_2,\\
    &\underline{\mathbf h}-\mathbf s_{i|k}^l
    \leq\mathbf h_{i|k}
    \leq\overline{\mathbf h}+\mathbf s_{i|k}^u,\\
    &\mathbf s_{i|k}^l,\mathbf s_{i|k}^u\geq0.
\end{aligned}
    \label{eq:residual_aware_nmpc_ocp}
\end{equation}
The dynamics and increment bounds apply for $i=0,\ldots,N-1$, the velocity
bounds for $i=1,\ldots,N$, and the soft bounds and nonnegative slacks for
$i=0,\ldots,N$. The parameter $a_{\max}$ limits the rate of change of the
commanded velocity, not the actual drone acceleration. When enabled,
$v_{\max}>0$ imposes hard bounds on both actual and commanded velocities;
the increment bound is always hard.

At each control step, the first optimized increment
$u_k=u_{0|k}^{\star}$ is selected as the control candidate,
where $\star$ denotes the optimizer's solution.
A nonzero solver status, a nonfinite increment, or a failed
hard increment-bound check produces a fallback candidate
$u_k=0$, corresponding to no change in the accumulated command.
Soft-constraint violations alone do not trigger fallback.
Accepted commands update the command accumulator and delay queue;
fallback events are logged and handled by the trial supervisor.

%=============================================================================
\section{Experiments}
\label{sec:experiments}
% Keep the paired tables and trajectory figure on separate text-page tops.

We first compare candidate models of the low-level actuation
response to assess the roles of command dead time and first-order
velocity dynamics in prediction.
This evaluation supports the actuation model used in the subsequent experiments.
We then address three research questions concerning tracking
control in simulation, residual-dynamics prediction, and physical
closed-loop performance:
\begin{itemize}
\item \textbf{RQ1:} Can nominal NMPC extend ball control from
constant-reference balancing to dynamic tracking, and improve tracking
accuracy over a nonlinear feedforward-feedback baseline in simulation?
\item \textbf{RQ2:} How does the model with the structured residual
compare with nominal dynamics, component ablations, and generic
learned residual models in causal multi-step prediction on
held-out real-world data?
\item \textbf{RQ3:} How does NMPC with the structured residual
compare with the baselines in accuracy, execution reliability,
and real-time feasibility during real-world constant-reference
balancing and dynamic tracking?
\end{itemize}

\begin{figure}[!tb]
    \centering
    \includegraphics[width=0.95\columnwidth]{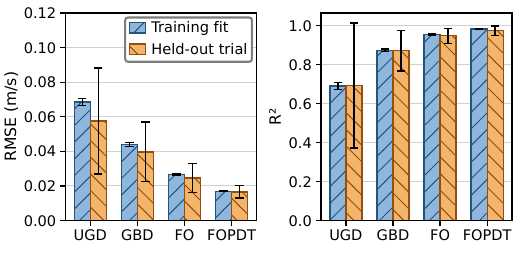}
    \caption{Comparison of four low-level vertical-velocity response models over 26 physical PRBS trials. Bars and error bars show means and sample standard deviations across trials, respectively.
    Training fit denotes a separate fit to each trial, while held-out trial denotes leave-one-trial-out prediction.}
    \label{fig:velocity_response_identification}
\end{figure}

\subsection{Identification of the Low-Level Actuation Response}
\label{sec:actuation_response_identification}
This experiment evaluates the FOPDT structure and identifies
its parameters for the low-level actuation response introduced
in Section ~\ref{sec:augmented_nominal_dynamics} \eqref{eq:fopdt_actuation_response}, relating commanded velocity
$c_z$ to actual velocity $v_{rz}$.
We used 26 physical pseudo-random binary sequence (PRBS) trials
collected at 60~Hz and compared four structures: unit gain with
delay (UGD), gain and bias with delay (GBD), first order without
delay (FO), and FOPDT. Each was fitted causally over complete response
sequences and evaluated by leave-one-trial-out validation, as shown in
Fig.~\ref{fig:velocity_response_identification}.
For FO and FOPDT, the velocity state is initialized with the first measured
velocity. Each model predicts the full response to the recorded commands
without subsequent correction from velocity measurements. The errors thus
assess complete response sequences rather than one-step prediction accuracy.
The pooled FOPDT parameters are
\begin{equation}
\begin{aligned}
    T_d &=0.0722~\mathrm{s}, & \tau&=0.1874~\mathrm{s},\\
    K   &=1.0135,           & b&=-0.00485~\mathrm{m/s}.
\end{aligned}
    \label{eq:identified_fopdt_parameters}
\end{equation}
FOPDT achieved the lowest mean held-out RMSE of $0.01659$~m/s, compared with $0.02459$~m/s for the next-best
FO model, $0.03971$~m/s for GBD, and $0.05754$~m/s for UGD. This ordering indicates that modeling the gradual
first-order response provides the main improvement over the pure-delay models, while retaining explicit dead time
yields a further gain. These results support the choice of FOPDT for the
augmented nominal dynamics.

\subsection{Tracking Control in Simulation}
\label{sec:simulation_tracking}
\textit{1) Simulation settings:}
We conduct the simulations in an environment developed using
Isaac Lab~\cite{2023_RAL_isaac_lab}, where a geometric SE(3)
controller~\cite{2010_SE3_controller} tracks the drone velocity
commands at 60~Hz.
We compare Nominal NMPC with a nonlinear feedforward-feedback (NFFB)
baseline. Nominal NMPC uses the augmented nominal dynamics without
residual correction in both its state predictor and optimization model.
Each controller is evaluated
on 500 episodes of 20~s for each reference type. Across
episodes, the environment samples $D_{\mathrm{sim}}\in\{6,7,8,9,10\}$,
$\tau_{\mathrm{sim}}\in[0.174,0.200]$~s, $K_{\mathrm{sim}}\in[0.95,1.05]$, and
$b_{\mathrm{sim}}\in[-0.0060,-0.0037]$~m/s. The subscript $\mathrm{sim}$ denotes simulated response parameters. Gaussian noise is also added
to the velocity response. Both
controllers use the fixed nominal values $D=8$,
$\tau=0.187$~s, $K=1.0$, and $b=-0.00485$~m/s. The simulated ball-beam mechanics follow the nominal model, with no
additional position-dependent or dissipative mismatch or external
disturbances. The prescribed uncertainty is confined to the low-level
actuation response. Nominal NMPC uses $N=30$, $h=1/60$~s, and the weights
\begin{equation}
\begin{aligned}
    \mathbf W&=\operatorname{diag}(20,4,0.1,0.2,0.2,0.05,0.25),\\
    \mathbf W_N&=\operatorname{diag}(100,20,0.5,1,1,0.05),
\end{aligned}
    \label{eq:nmpc_tracking_weights}
\end{equation}
with $\rho_1=\rho_2=10^4$. The soft bounds are
$\underline{\mathbf h}=[0,-40^\circ,-0.5]^\top$ and
$\overline{\mathbf h}=[0.70,40^\circ,0.5]^\top$, with position and tracking
error expressed in meters and angles converted to radians in the solver.
The command-rate limit is $a_{\max}=5$~m/s$^2$; hard velocity bounds are disabled.

The OCP is solved with acados~\cite{verschueren2022acados}
using sequential quadratic programming with real-time iteration
(SQP-RTI), a Gauss-Newton Hessian approximation, and the
discrete prediction map.
The resulting quadratic programs are solved using the
High-Performance Interior-Point Method (HPIPM)
solver~\cite{frison2020hpipm} with partial condensing.
The previous solution is shifted to warm-start each solve.

\textit{2) Reference design and metrics:}
We cover constant-reference balancing and dynamic tracking, with the latter represented by smooth B-spline and
transition-rich ramp-dwell trajectories:
\begin{itemize}
    \setlength{\itemsep}{0pt}
    \setlength{\parsep}{0pt}
    \setlength{\topsep}{2pt}
    \item \emph{Constant:} The goal and initial ball position are sampled independently from $[0.10,0.60]$~m,
    with at least $0.05$~m separation.
    \item \emph{Random B-spline:} A cubic 20-s trajectory uses 12 control points, starts and ends at $0.35$~m,
    and alternates its inner control points between $[0.10,0.30]$ and $[0.40,0.60]$~m to generate smooth motion.
    \item \emph{Random Ramp-dwell:} A 20-s reference starts at $0.35$~m and alternates six targets between the
    same intervals, with 4-5-s half-cycles drawn as either continuous ramps or ramp-and-dwell profiles to test
    slope changes, turning points, and dwell transitions.
\end{itemize}
The dynamic tracking initializes the ball on the reference. We evaluate the constant-reference balancing using steady-state error
(SE) and convergence time (CONT). SE is the mean absolute error over the final 1~s, while CONT is the time to
enter and remain within $0.01$~m of the target, with the episode duration assigned upon nonconvergence. Dynamic
tracking is evaluated by episode-level mean absolute error (MAE) and root
mean square error (RMSE), averaged across evaluated episodes. Results are reported as means and standard deviations across
evaluated episodes.

\begin{table}[!tb]
    \caption{Tracking control performance in simulation.}
    % Dynamic sample SDs: research/table_i_dynamic_dispersion/summary.csv (same runs as existing means).
    \label{tab:ideal_simulation_tracking}
    \centering
    {\footnotesize
    \setlength{\tabcolsep}{4pt}
    \renewcommand{\arraystretch}{1.25}
    \begin{tabular}{|c|c|c|c|}
        \hline
        \multirow{2}{*}{\textbf{Reference}} &
        \multirow{2}{*}{\textbf{Metric}} &
        \multicolumn{2}{c|}{\textbf{Method}} \\
        \cline{3-4}
        & & \textbf{NFFB} & \textbf{Nominal NMPC} \\
        \hline
        \multirow{2}{*}{\textbf{Constant}}
        & \textbf{SE} (cm) & $0.68{\pm}0.44$ & $\mathbf{0.27{\pm}0.16}$ \\
        \cline{2-4}
        & \textbf{CONT} (s) & $13.28{\pm}6.60$ & $\mathbf{3.73{\pm}0.79}$ \\
        \hline
        \multirow{2}{*}{\textbf{Random B-spline}}
        & \textbf{MAE} (cm) & $2.16{\pm}0.68$ & $\mathbf{0.79{\pm}0.26}$ \\
        \cline{2-4}
        & \textbf{RMSE} (cm) & $3.22{\pm}1.03$ & $\mathbf{1.68{\pm}0.64}$ \\
        \hline
        \multirow{2}{*}{\textbf{Random Ramp-dwell}}
        & \textbf{MAE} (cm) & $3.87{\pm}0.98$ & $\mathbf{0.64{\pm}0.47}$ \\
        \cline{2-4}
        & \textbf{RMSE} (cm) & $4.51{\pm}1.11$ & $\mathbf{1.13{\pm}0.75}$ \\
        \hline
    \end{tabular}}
\end{table}

\textit{3) Baseline:}
The NFFB controller uses the
same augmented nominal model and $D=8$ delay-compensation
predictor as Nominal NMPC, with time-aligned references.
Using the predicted states and omitting time indices, its
outer and inner feedback laws are
\begin{equation}
\begin{aligned}
    a_b^\star
    &=a_g-k_p(p_b-p_g)-k_v(v_b-v_g)-k_I z,\\
    \omega_c
    &=\omega_d-k_\theta(\theta-\theta_d).
\end{aligned}
    \label{eq:nffb_feedback}
\end{equation}
Here, $z$ integrates the position error with anti-windup,
and $a_g$ is obtained from consecutive preview velocities
and limited.
The desired ball acceleration $a_b^\star$ is clipped to
the range permitted by the beam-angle bounds and mapped
to a target angle by inverting
\eqref{eq:nominal_ball_acceleration}.
A second-order command filter with natural frequency
$\omega_f$ and damping ratio $\zeta_f$ generates the
bounded angle and rate targets $(\theta_d,\omega_d)$.
The angular-rate command $\omega_c$ is mapped to the desired
drone velocity $c_z^\star$ through \eqref{eq:beam_geometry};
the increment $c_z^\star-c_z(t_k^-)$ is then limited by
the command constraints to obtain $u_k$. 
The used gains are $k_p=0.80$~s$^{-2}$, $k_v=1.70$~s$^{-1}$,
$k_I=0.064$~s$^{-3}$, and $k_\theta=6$~s$^{-1}$.
The filter uses $\omega_f=14$~rad/s and $\zeta_f=1$,
with the integral state bounded by $|z|\leq0.25$~m\,s.
NFFB uses no residual model and requires no online
receding-horizon optimization.

\textit{4) Results:}
The results are summarized in Table~\ref{tab:ideal_simulation_tracking}. For constant-reference balancing, Nominal NMPC reduces SE and CONT by
$60.8\%$ and $71.9\%$, respectively. For B-spline references, it reduces MAE and RMSE by $63.7\%$ and $47.7\%$,
respectively. For ramp-dwell references, the corresponding reductions are $83.5\%$ and $74.9\%$. Nominal NMPC improves both constant-reference balancing and dynamic tracking,
with the largest relative error reductions on the ramp-dwell task. These
results support extending ball control to time-varying trajectories using
the nominal predictive control framework.

\subsection{Offline Residual-Dynamics Prediction}
\label{sec:offline_residual_prediction}

\begin{figure}[!tb]
    \centering
    \includegraphics[width=0.95\columnwidth]
    {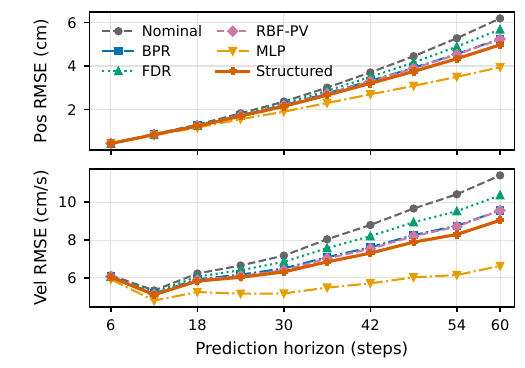}
    \caption{Held-out causal prediction errors of six dynamics
    models: ball-position (Pos) and ball-velocity (Vel)
    root-mean-square errors (RMSE) in the top and bottom panels,
    respectively. Horizons are measured in 60-Hz control steps.}
    \label{fig:offline_residual_rollout_rmse}
\end{figure}

\textit{1) Dataset:}
Real-world closed-loop data are processed as control snapshots
at 60~Hz and split by complete trials into 79 training,
6 validation, and 6 held-out test trials.
Each split contains constant, sine, triangle, and trapezoid
references, with no trial contributing windows to multiple splits.
Training and model selection use only the training and
validation sets; the test set is reserved for final evaluation.

\textit{2) Causal rollout protocol:}
We evaluate multi-step ball-position and ball-velocity prediction
to assess error accumulation over multiple horizons.
All models use identical initial states, pending-command queues,
recorded commands, FOPDT parameters, geometry, and RK4 integration.
Each rollout propagates the model for 60 steps without injecting
future measured states.
The residual is evaluated from the predicted state at each RK4 stage.
Endpoint position and velocity RMSEs are evaluated at
$H\in\{6,12,\ldots,60\}$ steps, corresponding to
$0.1$-$1.0$~s.

\textit{3) Models and training settings:}
We compare \textbf{Structured}
(Section~\ref{structured_residual_model}) with five alternatives:
\begin{itemize}
    \item Nominal Dynamics (\textbf{Nominal}): the augmented nominal dynamics in
    Section~\ref{sec:augmented_nominal_dynamics} with zero acceleration residual.
    \item Beam-Profile Residual (\textbf{BPR}): only the position-dependent beam-slope correction
    $r_{\mathrm{prof}}$ in \eqref{eq:profile_residual}, with 13 trained coefficients.
    \item Friction-Inspired Dissipative Residual (\textbf{FDR}): only
    $r_{\mathrm{diss}}$ in \eqref{eq:dissipative_residual}, with 26 trained
    coefficients and nonnegative Coulomb-like and viscous magnitudes.
    \item Position-Velocity Radial Basis Function Model (\textbf{RBF-PV}):
    representing the raw residual using a normalized $9\times9$ Gaussian basis
    over $(p_b,v_b)$. Position centers are uniformly distributed over the valid
    beam region, while velocity centers span $[-0.50,0.50]$~m/s with denser
    spacing near zero. The respective widths are $0.10$~m and $0.10$~m/s.
    \item Multilayer Perceptron (\textbf{MLP}): a $5$-$16$-$16$-$1$ network with tanh hidden layers
    approximating the raw ball-acceleration residual. Its inputs
    $(p_b,v_b,\theta,\omega,a_b^{\mathrm{nom}})$ are standardized using training-set statistics only.
\end{itemize}
Structured uses $M=13$ position bases,
$\sigma_p=0.065$~m, $\epsilon=10^{-12}$, and
$v_0=0.025$~m/s, with coefficient bounds
$\bar\delta_\theta=0.035$~rad,
$\bar f_c=0.15$~m/s$^2$, and
$\bar f_v=0.50$~s$^{-1}$.
All learned models share the output bound
$\bar r_a=0.25$~m/s$^2$.
BPR, FDR, and RBF-PV use
$r_a=\bar r_a\tanh(r_a^{\mathrm{raw}}/\bar r_a)$,
whereas MLP uses $r_a=\bar r_a\tanh(z)$ with dimensionless
linear output $z$.

All five residual models are trained independently from scratch
using the data-loss terms in
\eqref{eq:structured_residual_training_loss}, with
$\lambda_{\mathrm{pt}}=0.25$ and
$\lambda_{\mathrm{roll}}=0$ and $0.5$ in Stages 1 and 2,
respectively.
Stage 2 starts from the best Stage 1 checkpoint.
For this six-model comparison, training uses the same Adam,
window-sampling protocol, batch size, and seed, with
deterministic CPU float64 computation.
Stages 1/2 use at most 80/50 epochs, 16/8 updates per epoch,
and learning rates of $0.003/0.0003$.
The integral horizons are
$\mathcal H_{\mathrm{int}}=\{3,6,12\}$ with weights
$\alpha_H=(0.50,0.30,0.20)$; the rollout horizons are
$\mathcal H_{\mathrm{roll}}=\{6,12,30,33\}$ with weights
$\beta_H=(0.35,0.30,0.20,0.15)$.
The remaining loss settings are
$\delta_a=0.02$~m/s$^2$, $\delta_r=1$,
$s_p=0.01$~m, $s_v=0.05$~m/s, and $\eta_c=0.25$.

In both stages, Structured uses the smoothness penalty in
\eqref{eq:rbf_smoothness_loss} with
$\lambda_{\mathrm{sm}}=1$, $\eta_1=0.002$, and $\eta_2=0.01$;
BPR and FDR retain their respective component penalties.
RBF-PV penalizes mean squared first and second differences
of coefficients normalized by $\bar r_a$ along both grid axes,
with weights $0.002$ and $0.01$, respectively.
MLP applies an L2 penalty to its weights and biases
with coefficient $10^{-5}$.

\begin{table}[!tb]
    \caption{Training-objective ablation for multi-step prediction.}
    \label{tab:training_objective_ablation}
    \centering
    {\footnotesize
    \setlength{\tabcolsep}{3pt}
    \renewcommand{\arraystretch}{1.15}
    \begin{tabular}{|l|cc|cc|}
        \hline
        \multirow{2}{*}{\textbf{Objective}} &
        \multicolumn{2}{c|}{\boldmath$H=30$} & \multicolumn{2}{c|}{\boldmath$H=60$} \\
        \cline{2-5}
        & \textbf{Pos} (cm) & \textbf{Vel} (cm/s) &
        \textbf{Pos} (cm) & \textbf{Vel} (cm/s) \\
        \hline
        \multicolumn{5}{|c|}{\textbf{Structured}} \\
        \hline
        \textbf{Full} & \mbox{\boldmath$2.14{\pm}0.00$} & \mbox{\boldmath$6.32{\pm}0.01$} & \mbox{\boldmath$4.98{\pm}0.02$} & \mbox{\boldmath$9.05{\pm}0.03$} \\
        \textbf{w/o rollout} & $2.27{\pm}0.01$ & $6.69{\pm}0.04$ & $5.61{\pm}0.05$ & $10.17{\pm}0.09$ \\
        \textbf{w/o integral} & $2.15{\pm}0.00$ & $6.33{\pm}0.01$ & $4.99{\pm}0.02$ & $9.07{\pm}0.03$ \\
        \textbf{w/o both} & $2.18{\pm}0.00$ & $6.41{\pm}0.02$ & $5.12{\pm}0.01$ & $9.35{\pm}0.03$ \\
        \hline
        \multicolumn{5}{|c|}{\textbf{MLP}} \\
        \hline
        \textbf{Full} & \mbox{\boldmath$1.88{\pm}0.02$} & \mbox{\boldmath$5.12{\pm}0.05$} & \mbox{\boldmath$3.89{\pm}0.06$} & $6.55{\pm}0.10$ \\
        \textbf{w/o rollout} & $1.97{\pm}0.01$ & $5.39{\pm}0.01$ & $4.28{\pm}0.03$ & $7.29{\pm}0.08$ \\
        \textbf{w/o integral} & $1.90{\pm}0.02$ & $5.20{\pm}0.08$ & $3.94{\pm}0.09$ & \mbox{\boldmath$6.49{\pm}0.17$} \\
        \textbf{w/o both} & $2.01{\pm}0.02$ & $5.51{\pm}0.07$ & $4.37{\pm}0.09$ & $7.30{\pm}0.17$ \\
        \hline
    \end{tabular}
    }
\end{table}

\textit{4) Results:}
Fig.~\ref{fig:offline_residual_rollout_rmse} compares the
held-out rollout RMSEs across prediction horizons.
Structured improves both metrics over Nominal at every
evaluated horizon, reducing position/velocity RMSE by
$9.3\%/12.0\%$ at $H=30$ and $19.7\%/20.8\%$ at $H=60$.
It also outperforms BPR, FDR, and RBF-PV in these pooled metrics,
supporting the combined profile and dissipative corrections.
MLP achieves the lowest offline errors, with corresponding
reductions over Nominal of $19.7\%/27.9\%$ at $H=30$ and
$36.3\%/41.9\%$ at $H=60$. However, lower offline prediction error does not necessarily
translate into better closed-loop control, as examined in
Section~\ref{sec:real_world_closed_loop_evaluation}.

\textit{5) Training-objective ablation:}
For Structured and MLP, we ablate the training objective in
\eqref{eq:structured_residual_training_loss} by removing the
rollout term, the integral-consistency term, or both.
Each configuration is trained from scratch with three seeds, 
retaining the same model structure, regularization,
two-stage training schedule, and validation-based selection
criterion. Table~\ref{tab:training_objective_ablation}
reports prediction RMSEs at $H=30$ and $H=60$, summarized
across seeds after pooling test-window errors within each seed.

Removing the rollout term increases position and velocity
RMSEs at both horizons for both models.
Removing the integral term has a smaller effect: Structured
remains close to Full, while MLP obtains a slightly lower mean
velocity RMSE at $H=60$.
For Structured, removing both terms yields lower errors than
removing rollout alone, indicating that the effect of integral
consistency depends on the retained training terms.
These results support causal rollout training, whereas the
additional benefit of integral consistency is modest and
configuration-dependent.

\begin{figure}[!tb]
    \centering
    \includegraphics[width=0.73\columnwidth,height=0.42\columnwidth]{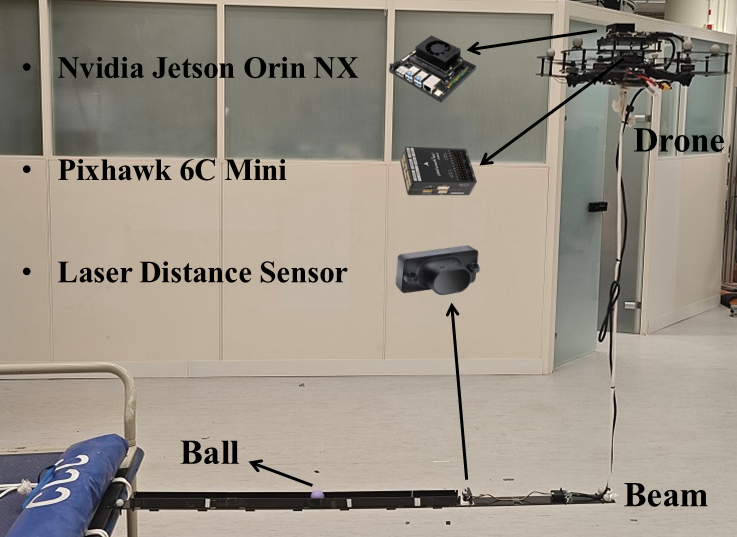}
    \caption{Physical drone-ball platform. The drone carries an NVIDIA Jetson Orin NX and a Pixhawk 6C Mini
    and is connected to the beam by a rope. A laser distance sensor measures the ball position along the beam.}
    \label{fig:real_world_system}
\end{figure}

\subsection{Real-World Closed-Loop Evaluation}
\label{sec:real_world_closed_loop_evaluation}

\begin{table*}[t]
    \centering
    \footnotesize
    \caption{Real-world constant-reference balancing performance.}
    \label{tab:real_world_constant_tracking}
    \centering
    {\footnotesize
    \setlength{\tabcolsep}{4pt}
    \renewcommand{\arraystretch}{1.25}
    \begin{tabular}{|c|ccc|ccc|ccc|}
        \hline
        \multirow{2}{*}{\textbf{Target}} & \multicolumn{3}{c|}{\textbf{SE} (cm)} & \multicolumn{3}{c|}{\textbf{CONT} (s)} & \multicolumn{3}{c|}{\textbf{Completed/Attempted}} \\
        \cline{2-10}
        & \textbf{Nom.} & \textbf{MLP} & \textbf{Str.} & \textbf{Nom.} & \textbf{MLP} & \textbf{Str.} & \textbf{Nom.} & \textbf{MLP} & \textbf{Str.} \\
        \hline
        \textbf{0.15 m} & \mbox{$3.70{\pm}0.80$} & \mbox{$25.40{\pm}16.64$} & \mbox{\boldmath$2.13{\pm}0.95$} & \mbox{$25.6{\pm}10.9$} & \mbox{$30.0{\pm}0.0$} & \mbox{\boldmath$8.5{\pm}10.7$} & 6/6 & 6/6 & 6/6 \\
        \hline
        \textbf{0.35 m} & \mbox{$1.02{\pm}0.46$} & \mbox{$14.29{\pm}9.82$} & \mbox{\boldmath$0.85{\pm}0.80$} & \mbox{$5.1{\pm}6.3$} & \mbox{$30.0{\pm}0.0$} & \mbox{\boldmath$4.6{\pm}5.3$} & 6/6 & 6/6 & 6/6 \\
        \hline
        \textbf{0.55 m} & \mbox{$6.21{\pm}2.00$} & \mbox{$18.66{\pm}8.68$} & \mbox{\boldmath$2.41{\pm}1.81$} & \mbox{$30.0{\pm}0.0$} & \mbox{$30.0{\pm}0.0$} & \mbox{\boldmath$16.2{\pm}15.2$} & 6/6 & 6/6 & 6/6 \\
        \hline
        \textbf{Overall} & \mbox{$3.64$} & \mbox{$19.45$} & \mbox{\boldmath$1.80$} & \mbox{$20.2$} & \mbox{$30.0$} & \mbox{\boldmath$9.8$} & 18/18 & 18/18 & 18/18 \\
        \hline
    \end{tabular}}
\par\vspace{10pt}
    \caption{Real-world dynamic tracking performance.}
    \label{tab:real_world_dynamic_tracking}
    \centering
    {\footnotesize
    \setlength{\tabcolsep}{4pt}
    \renewcommand{\arraystretch}{1.25}
    \resizebox{\textwidth}{!}{%
    \begin{tabular}{|c|ccc|ccc|ccc|ccc|}
        \hline
        \multirow{2}{*}{\textbf{Reference}} & \multicolumn{3}{c|}{\textbf{MAE} (cm)} & \multicolumn{3}{c|}{\textbf{RMSE} (cm)} & \multicolumn{3}{c|}{\textbf{G}} & \multicolumn{3}{c|}{\textbf{Completed/Attempted}} \\
        \cline{2-13}
        & \textbf{Nom.} & \textbf{MLP} & \textbf{Str.} & \textbf{Nom.} & \textbf{MLP} & \textbf{Str.} & \textbf{Nom.} & \textbf{MLP} & \textbf{Str.} & \textbf{Nom.} & \textbf{MLP} & \textbf{Str.} \\
        \hline
        \textbf{Sine 8 s} & \mbox{$2.57{\pm}0.34$} & \mbox{$12.11{\pm}2.34$} & \mbox{\boldmath$1.33{\pm}0.27$} & \mbox{$3.08{\pm}0.37$} & \mbox{$16.50{\pm}2.90$} & \mbox{\boldmath$1.95{\pm}0.67$} & \mbox{$0.825{\pm}0.025$} & \mbox{$0.627{\pm}0.120$} & \mbox{\boldmath$0.949{\pm}0.018$} & 6/6 & 4/6 & 6/6 \\
        \hline
        \textbf{Sine 6 s} & \mbox{$2.33{\pm}0.33$} & \mbox{$15.34{\pm}8.13$} & \mbox{\boldmath$1.32{\pm}0.23$} & \mbox{$3.15{\pm}0.36$} & \mbox{$18.09{\pm}8.57$} & \mbox{\boldmath$2.25{\pm}0.39$} & \mbox{$0.842{\pm}0.026$} & \mbox{$0.576{\pm}0.280$} & \mbox{\boldmath$0.951{\pm}0.023$} & 6/6 & 6/6 & 6/6 \\
        \hline
        \textbf{Triangle 8 s} & \mbox{$2.35{\pm}0.31$} & \mbox{$17.14{\pm}5.58$} & \mbox{\boldmath$1.30{\pm}0.18$} & \mbox{$3.01{\pm}0.63$} & \mbox{$20.59{\pm}4.80$} & \mbox{\boldmath$1.76{\pm}0.32$} & \mbox{$0.819{\pm}0.028$} & \mbox{$0.380{\pm}0.219$} & \mbox{\boldmath$0.937{\pm}0.009$} & 6/6 & 5/6 & 6/6 \\
        \hline
        \textbf{Triangle 6 s} & \mbox{$2.26{\pm}0.16$} & \mbox{$10.08{\pm}8.01$} & \mbox{\boldmath$1.61{\pm}0.20$} & \mbox{$2.91{\pm}0.21$} & \mbox{$12.47{\pm}8.45$} & \mbox{\boldmath$2.38{\pm}0.52$} & \mbox{$0.837{\pm}0.018$} & \mbox{$0.751{\pm}0.301$} & \mbox{\boldmath$0.920{\pm}0.023$} & 6/6 & 6/6 & 6/6 \\
        \hline
        \textbf{Trapezoid 8 s} & \mbox{$2.83{\pm}0.14$} & \mbox{$16.60{\pm}6.71$} & \mbox{\boldmath$1.70{\pm}0.20$} & \mbox{$3.61{\pm}0.19$} & \mbox{$19.56{\pm}6.75$} & \mbox{\boldmath$2.38{\pm}0.21$} & \mbox{$0.835{\pm}0.015$} & \mbox{$0.541{\pm}0.233$} & \mbox{\boldmath$0.956{\pm}0.014$} & 6/6 & 5/6 & 6/6 \\
        \hline
        \textbf{Trapezoid 6 s} & \mbox{$2.73{\pm}0.24$} & \mbox{$5.39{\pm}3.24$} & \mbox{\boldmath$2.20{\pm}0.21$} & \mbox{$3.75{\pm}0.27$} & \mbox{$6.45{\pm}3.24$} & \mbox{\boldmath$3.38{\pm}0.47$} & \mbox{$0.867{\pm}0.010$} & \mbox{\boldmath$1.043{\pm}0.080$} & \mbox{$0.941{\pm}0.015$} & 6/6 & 4/6 & 6/6 \\
        \hline
        \textbf{Overall} & \mbox{$2.51$} & \mbox{$12.78$} & \mbox{\boldmath$1.58$} & \mbox{$3.25$} & \mbox{$15.61$} & \mbox{\boldmath$2.35$} & \mbox{$0.838$} & \mbox{$0.653$} & \mbox{\boldmath$0.942$} & 36/36 & 30/36 & 36/36 \\
        \hline
    
\end{tabular}}}
\end{table*}

\textit{1) Experimental setup:}
We evaluate the controllers on the physical platform shown in
Fig.~\ref{fig:real_world_system}. The drone is equipped with an NVIDIA Jetson Orin NX onboard computer and a
Pixhawk 6C Mini flight controller running PX4~\cite{2015_px4} for low-level velocity control.
A rope connects the drone to one end of the beam, while the opposite end
is supported at a stationary pivot. Vertical drone motion changes the beam inclination and guides the freely
rolling ball, whose position is measured by a laser distance sensor.
To evaluate closed-loop performance across diverse reference motions, we use three constant reference positions,
$p_g^{\mathrm{const}}\in\{0.15,0.35,0.55\}$~m, together with sine, triangle, and trapezoid references with periods of 8 and 6~s.
The constant targets span the beam center and two off-center regions, while the dynamic references vary in speed
and transition sharpness. Residual parameters remain fixed throughout the tests.

\textit{2) Evaluation metrics:}
Only normally completed 30-s trials enter the tracking-error statistics. For constant-reference balancing, we retain the SE and CONT definitions used in the ideal-simulation
evaluation, but use a 5-s terminal window for SE and a $0.03$-m convergence threshold for CONT; a nonconverged
trial is assigned 30~s. Dynamic tracking is evaluated over
the complete trial using position MAE, RMSE, and response gain $G=A_b/A_g$.
For each controller, $A_b$ and $A_g$ are obtained by independently fitting the ball and reference positions
to a constant plus sine and cosine terms at the known reference frequency over the full trial.
Thus, $G$ measures the fundamental-component gain, including for triangle and trapezoid references.
Each condition is averaged over its actual number of complete trials, 
and the overall result equally weights the condition means.

\textit{3) Baselines:}
We compare the proposed structured-residual-aware NMPC (\textbf{Structured-NMPC}) with two baselines:
\begin{itemize}
    \item Nominal NMPC (\textbf{Nominal-NMPC}): uses the augmented nominal dynamics in both the
    delay-aware state predictor and NMPC, with zero acceleration residual.
    \item MLP-residual-aware NMPC (\textbf{MLP-NMPC}): uses the same frozen five-input MLP acceleration
    residual in both the predictor and NMPC.
\end{itemize}
All three controllers use the same solver configuration
and warm-start procedure as in
Section~\ref{sec:simulation_tracking}.
Per-trial configuration checks confirm common state information, objective weights, constraints,
$N=30$, 60-Hz operation, identified FOPDT parameters, $D=4$, and deployment interfaces.
The physical controllers use the weighting matrices in
\eqref{eq:nmpc_tracking_weights}, $\rho_1=\rho_2=10^4$,
$a_{\max}=5$~m/s$^2$, and enabled velocity bounds $v_{\max}=3$~m/s.
The trial configurations specify soft bounds
$\underline{\mathbf h}=[0,-40^\circ,-0.7]^\top$ and
$\overline{\mathbf h}=[0.75,40^\circ,0.7]^\top$, with position and tracking
error in meters and angles converted to radians.
A solver fallback terminates the physical trial without publishing the
rejected candidate command.
The prediction-model difference is the acceleration residual: zero, MLP, or structured.
The MLP-NMPC tests use the same frozen MLP checkpoint as the offline prediction evaluation.

\begin{figure*}[t]
    \centering
    \includegraphics[width=0.88\textwidth]{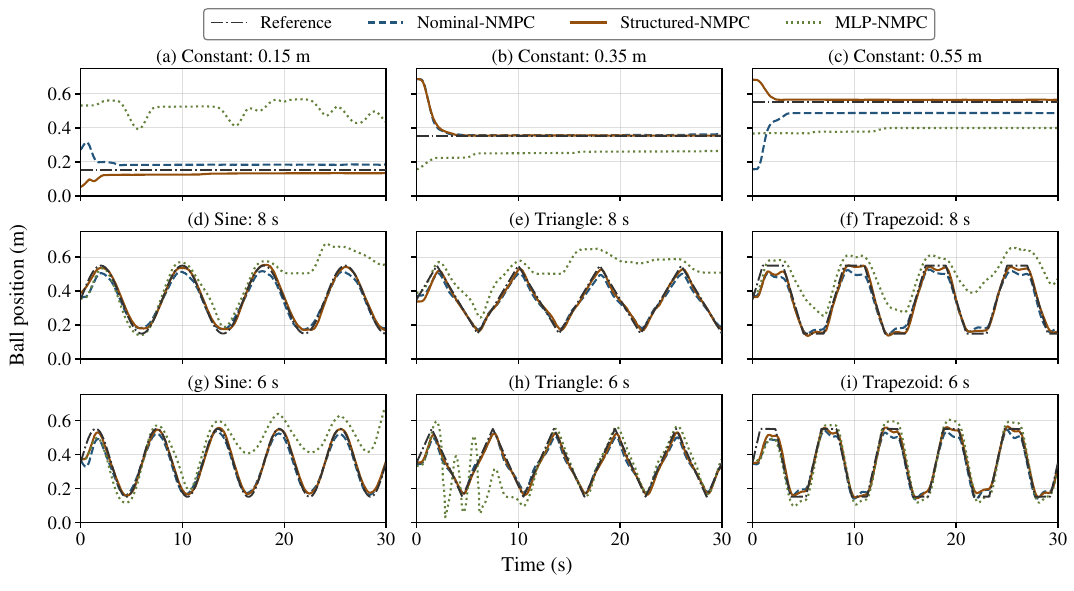}
    \caption{Representative real-world tracking trials for three controllers.
    Among normally completed trials, traces use the middle trial ranked by SE (constant) or MAE (dynamic).
    The rows show constant targets, 8-s references, and 6-s references, respectively.}
    \label{fig:real_world_closed_loop_tracking}
\end{figure*}

\textit{4) Tracking results:}
Fig.~\ref{fig:real_world_closed_loop_tracking} compares the reference and measured ball-position trajectories
of the three controllers for three constant targets and six dynamic references with periods of 8 and 6~s.
Table~\ref{tab:real_world_constant_tracking} reports SE and CONT
for constant-reference balancing, while
Table~\ref{tab:real_world_dynamic_tracking} reports MAE, RMSE,
and response gain for dynamic tracking at 8- and 6-s periods.
The figure shows individual representative trials, whereas
the tables summarize normally completed trials as mean $\pm$
sample standard deviation for each condition.
Nom., MLP, and Str. denote Nominal-NMPC, MLP-NMPC, and Structured-NMPC, respectively.
For constant targets, Structured-NMPC reduces overall SE by $50.7\%$ and CONT by $51.8\%$ relative to Nominal-NMPC.
Its largest SE reduction, $61.2\%$, occurs at $0.55$~m, while center-target SE decreases by $16.8\%$.
Across the six dynamic conditions, MAE and RMSE decrease by $37.2\%$ and $27.8\%$.
The overall response gain increases from $0.838$ to $0.942$, reducing its deviation from unity by
$64.6\%$.
Consistent with these aggregate results, Structured-NMPC achieves
lower MAE, RMSE, and gain deviation than Nominal-NMPC across
all six tested dynamic conditions.

MLP-NMPC exhibits larger overall tracking errors and less reliable execution:
SE and CONT for constant-reference balancing are $19.449$~cm and $30.000$~s, respectively, compared with
$1.796$~cm and $9.751$~s for Structured-NMPC; over all six dynamic conditions,
MAE is $12.777$ versus $1.576$~cm and RMSE is $15.611$ versus $2.349$~cm.
A near-unity gain alone does not ensure low tracking error: the four completed MLP 6-s trapezoid trials
have a mean $G=1.043$ but MAE of $5.391$~cm.
Within the reported set, MLP-NMPC completes all 18 constant-reference trials and 30 of 36 dynamic trials,
with four action-bound-check fallback terminations,
one beam-angle safety termination, and one manual hover intervention following large drone oscillations.
All four reported fallback events have solver status zero but fail the deployment action-bound check.
While Nominal-NMPC and Structured-NMPC complete all 54 reported attempts.

\textit{5) Real-time feasibility:}
Timing statistics use all available computation-time records across all reported attempts.
Nominal-NMPC has a mean computation time of $1.594$~ms, a 99th-percentile time of $2.738$~ms,
and a maximum of $3.452$~ms. Structured-NMPC has a mean of $1.767$~ms, a 99th percentile of
$2.984$~ms, and a maximum of $3.731$~ms. MLP-NMPC has a mean of $2.421$~ms,
a 99th percentile of $4.252$~ms, and a maximum of $11.094$~ms.
All recorded computation times are below the $16.67$-ms control period, and no scheduled soft or hard
deadline misses are logged, supporting the computational feasibility of 60-Hz operation on the tested platform.

\section{Discussion: Prediction Accuracy versus Closed-Loop Performance}
\label{sec:prediction_control_discussion}
The offline results in Fig.~\ref{fig:offline_residual_rollout_rmse} and
the real-world results in Tables~\ref{tab:real_world_constant_tracking}
and~\ref{tab:real_world_dynamic_tracking} reveal a discrepancy: MLP predicts more accurately offline, whereas
Structured-NMPC achieves better aggregate tracking performance. 
Offline rollouts test whether
the model reproduces motion under recorded commands, whereas NMPC uses the model to optimize
commands for tracking.
This predictive-control decision also depends on local sensitivities: how predicted 
motion changes with state or input perturbations. A small average prediction error does
not by itself establish that the model accurately captures these local sensitivities. Gupta et al.~\cite{gupta2020structured} reported
a related finding in a simulated cart-pole task: a black-box model
predicted slightly more accurately but performed worse in
control than a structured model, whose derivatives better
matched the effects of state and input changes.
This motivates examining how the residual models alter these sensitivities and
the predicted command-to-position responses over the NMPC horizon.

\subsection{Local Sensitivities of the Prediction Models}
\label{subsec:discussion-sensitivities}
The residual models use different representations. Comparing their local
sensitivities helps identify how these representations change the responses
used in prediction.

To quantify these differences, we evaluate the Nominal, MLP, and Structured
models at the same states collected from MLP-NMPC and Structured-NMPC trials.
We estimate the sensitivity of the total predicted ball acceleration
$a_b^{\mathcal{F}}$ for model $\mathcal{F}$ using central differences, perturbing one
state at a time while holding the others fixed and recomputing dependent
model inputs. The position sensitivities of MLP and Structured have similar
negative medians, although their distributions differ. Ball-velocity
sensitivities are nonpositive at all sampled states for Structured and
negative at almost all sampled states for MLP. Thus, the MLP also learns
predominantly negative ball-velocity dependence without enforcing it
structurally. These shared signs do not imply identical dynamics; we next
examine the beam-angle and drone-velocity responses.

We quantify the beam-angle response relative to Nominal by
$\rho_\theta=(\partial a_b^{\mathcal{F}}/\partial\theta)/
(\partial a_b^{\mathrm{nom}}/\partial\theta)$.
The median ratio is 0.808 for MLP and
1.000 for Structured. Thus, MLP generally attenuates the
predicted beam-angle response over the sampled states, whereas Structured
largely preserves it. Beam-angle sensitivity alone does not describe the full effect of a
velocity command, which acts through the drone's vertical velocity
$v_{rz}$. We therefore also examine how changes in $v_{rz}$ affect the
predicted ball acceleration at a fixed beam angle.
From \eqref{eq:beam_geometry}, $\omega=\kappa(\theta)v_{rz}$, where
$\kappa(\theta)$ is the geometric conversion factor; perturbing $v_{rz}$
therefore also changes $\omega$. In the nominal model, this contribution
to ball acceleration is quadratic in $\omega$, so its first-order
sensitivity to $v_{rz}$ vanishes at rest. Structured preserves this
property because its residual does not depend on $v_{rz}$ or $\omega$.
By contrast, the MLP's explicit $\omega$ input allows an additional
first-order response even at zero drone velocity. We quantify the
difference from the nominal velocity sensitivity as
\begin{equation}
 \Delta J_{v_{rz}}^{\mathcal{F}}
 =\frac{\partial a_b^{\mathcal{F}}}{\partial v_{rz}}
  -\frac{\partial a_b^{\mathrm{nom}}}{\partial v_{rz}}.
 \label{eq:local-velocity-sensitivity}
\end{equation}
As shown in Fig.~\ref{fig:local-model-response}(a), the median is
$-0.691$~s$^{-1}$ for MLP, with an interquartile range of
$[-0.776,-0.622]$~s$^{-1}$, whereas Structured introduces
no additional velocity sensitivity. For the same small increase in
drone velocity, MLP therefore shifts the predicted acceleration change
in the negative ball-coordinate direction relative to Nominal.
This instantaneous difference must be considered together with the
evolving beam angle to understand the predicted motion over the
control horizon.

\subsection{Predicted Responses over the Control Horizon}
\label{subsec:discussion-responses}
We examine the predicted ball-position response to a small change in
the velocity command. Each model is initialized at its own static
equilibrium for the 0.35-m target: $p_b=0.35$~m, $v_b=v_{rz}=0$, and
the beam angle that makes its predicted ball acceleration zero. The command $c_{\mathrm{eq}}=-b/K$ balances the low-level response
bias. Using $\epsilon=10^{-4}$~m/s, we apply commands $c_{\mathrm{eq}}+\epsilon$ and
$c_{\mathrm{eq}}-\epsilon$ in separate rollouts, hold each command constant, and calculate
\begin{equation}
 S_p^{\mathcal{F}}(t)=\frac{p_b^{\mathcal{F}}(t;c_{\mathrm{eq}}+\epsilon)
                    -p_b^{\mathcal{F}}(t;c_{\mathrm{eq}}-\epsilon)}{2\epsilon}.
 \label{eq:local-command-response}
\end{equation}
Here, $S_p^{\mathcal{F}}(t)$ measures the position change per unit change in commanded velocity. Its sign indicates the direction of the local position
response. The rollouts use the deployed integration scheme and low-level velocity
dynamics. Time starts after delay compensation, so command dead time is not applied
again.

\begin{figure}[t]
    \centering
    \includegraphics[width=\columnwidth]{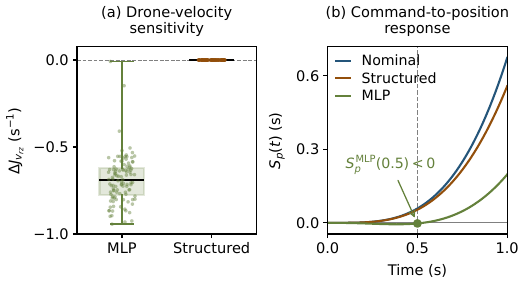}
    \caption{Local responses of the frozen prediction models. (a) Additional
    drone-velocity sensitivity relative to Nominal, as defined in
    \eqref{eq:local-velocity-sensitivity}. Each point is the median for one trial.
    (b) Position sensitivity to a held velocity-command perturbation, starting from
    each model's own equilibrium at 0.35~m.
    }
    \label{fig:local-model-response}
\end{figure}

Fig.~\ref{fig:local-model-response}(b) shows the resulting position
sensitivities $S_p^{\mathcal{F}}(t)$. At the $0.5$-s NMPC horizon, their values are
$-0.00241$~s for MLP, $0.0585$~s for Nominal, and
$0.0534$~s for Structured. MLP therefore predicts a small
opposite-direction displacement at this horizon, although its response becomes
positive later. This behavior is consistent with competing contributions from drone
velocity and beam tilt. Near the horizontal configuration, positive drone velocity
reduces $\theta$, which increases the nominal gravity-driven ball acceleration.
MLP's additional negative velocity sensitivity can oppose that contribution before
the beam tilt has accumulated. The resulting short-time cancellation changes the
predicted displacement available within the control horizon, even when prediction
under recorded commands is accurate.

To check whether the sign difference is specific to the separate
equilibrium initializations, we also apply identical command perturbations
to all three models from common states sampled from the recorded trials.
The results show that, at some sampled states, MLP predicts an opposite
position response to Nominal and Structured at the NMPC horizon, even
under the same initial state and command perturbation.
Such response differences can alter the predicted tracking benefit
and its tradeoff with control penalties.

\subsection{Implications and Limitations}
\label{subsec:discussion-implications}
The local-response analysis shows that the MLP residual alters beam-angle
and drone-velocity sensitivities and can change the direction of the predicted
command-to-position response within the NMPC horizon. These changes may
contribute to MLP-NMPC's poor closed-loop tracking by altering the predicted
benefits of candidate commands. Physical structure provides a compact way
to encode prior constraints on the responses used by predictive control.
However, these comparisons do not establish
which model more accurately captures the physical system's local responses.
We also examine three additional possible
explanations: solver convergence, static-equilibrium mismatch, and
training-deployment distribution shift, in the Supplementary Material,
Section I. These analyses provide possible explanations for the
observed degradation, but do not establish a complete causal account or
quantify the relative contributions of the proposed mechanisms.
These findings therefore motivate evaluating residual
models through prediction accuracy, the responses used in control decisions,
and actual closed-loop performance.

%=============================================================================
\section{Conclusion and Future Work}
\label{sec:conclusions}
This paper developed a predictive control framework with structured
residual dynamics for indirect drone-ball manipulation, extending
constant-reference balancing to dynamic tracking.
The framework augments ball-beam mechanics with an identified FOPDT
actuation model, compensates for command dead time through state
prediction, and incorporates a learned structured ball-acceleration residual
into both the predictor and NMPC.
Simulations show lower tracking errors than NFFB.
The structured residual improves multi-step prediction over nominal
dynamics on real data and aggregate physical tracking
accuracy over nominal NMPC, while retaining real-time operation.
Although MLP predicts more accurately offline, Structured-NMPC achieves
better aggregate closed-loop accuracy and more reliable execution,
supporting the value of structured residual dynamics for  closed-loop control.

The framework is evaluated on the vertically constrained
Aerial-Balance-Bench system. Future work will explore broader
operating conditions with less constrained drone motion and a moving
support point at the opposite end of the beam.

\bibliographystyle{IEEEtran}
\bibliography{references}

\end{document}

\maketitle

\section{Additional Diagnostics of MLP-NMPC Closed-Loop Performance}
\label{sec:mlp-diagnostics}
As presented in the main paper, the MLP residual model achieves the lowest offline prediction errors among the evaluated models. However, MLP-NMPC performs poorly in the real-world closed-loop experiments, with larger overall tracking errors than Nominal-NMPC.
In contrast, the proposed structured residual model improves offline prediction accuracy relative to the nominal model, and the resulting Structured-NMPC achieves the best overall closed-loop tracking performance among the evaluated controllers. These contrasting outcomes raise a natural question: what factors underlie the poor closed-loop performance of MLP-NMPC and the superior performance of Structured-NMPC?
Section~VI, ``Discussion: Prediction Accuracy versus Closed-Loop Performance,'' of the main paper examines a possible explanation based on altered model responses: the MLP residual changes sensitivities to beam angle and drone velocity and can modify the predicted command-to-position response within the NMPC horizon. These changes may influence NMPC's control decisions and contribute to poor closed-loop tracking. This section complements that discussion by considering other possible explanations, providing a more comprehensive assessment of the observed performance degradation of MLP-NMPC.
Three additional hypotheses are examined:
\begin{itemize}
    \item \textbf{Online Solver Convergence.}
    The deployed solver settings and limited iteration budget may produce insufficiently converged solutions to the online optimal control problem (OCP) formulated by MLP-NMPC, resulting in control actions that degrade closed-loop tracking.
    
    \item \textbf{Static-Equilibrium Consistency.}
    The MLP residual may misrepresent the acceleration balance near stationary operating conditions, leading to inaccurate predictions of the beam inclination required to hold the ball and contributing to persistent constant-reference tracking errors.

    \item \textbf{Input Distribution Shift.}
    The input feature distribution encountered by the MLP residual model during closed-loop operation may differ from its offline training distribution. In particular, visits to regions poorly covered by the training data may reduce residual prediction accuracy and degrade closed-loop tracking.
\end{itemize}

\subsection{Online Solver Convergence}
\label{subsec:solver}
\subsubsection{Solver Behavior of the Original MLP-NMPC}
\begin{figure}[tb]
\centering
\includegraphics[width=0.98\linewidth]{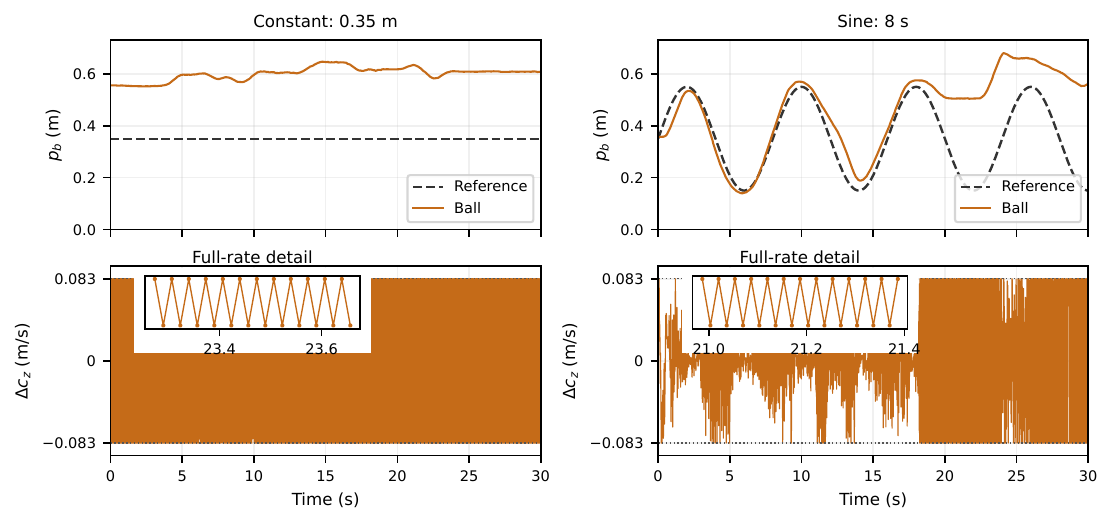}
\caption{Original MLP-NMPC trials illustrating sustained full-bound oscillation.}
\label{fig:solver-original}
\end{figure}

As described in Section~V-B of the main paper, we use sequential
quadratic programming with real-time iteration (SQP-RTI) to solve the OCP of NMPC. Under this scheme, we perform a single SQP update per control step to meet real-time computational requirements. We simply set the Levenberg-Marquardt (LM) regularization parameter to $0.0$, although this regularization can improve numerical conditioning by adding a diagonal term to the Hessian approximation.
Inspection of the MLP-NMPC experimental results reveals sustained full-bound oscillations in the velocity-command increments ($\Delta c_z$) in many trials. Fig.~\ref{fig:solver-original} shows two illustrative examples for a 0.35-m constant target and an 8-s sine reference. During these oscillations, successive increments alternate between opposite bounds $\pm U$, with $U=0.083$ m/s, producing little net change in the accumulated velocity command and limiting the sustained command correction. 

To further assess convergence of the original MLP-NMPC solver, we examine \RevBaselineSnapshots{} frozen OCPs spanning all \RevBaselineTrials{} MLP-NMPC trails reported in the main paper. Snapshots are taken at the recorded control steps closest to 5, 10, and 20 s. For records ending before 5 s, an early control step near 0.5 s and the final valid control step are used instead. Selection is independent of tracking error. The delay-predicted initial state, previous command, reference preview, and OCP parameters are held fixed. Starting from optimizer memory reconstructed by recorded-input RTI replay, the original update is repeated without shifting the horizon, for at most 2000 updates including the first update at the selected control step. We assess numerical convergence using an aggregate residual metric $R$, defined as the maximum of four residual components:
\begin{equation}
 R=\max\{R_{\mathrm{stat}},R_{\mathrm{eq}},R_{\mathrm{ineq}},R_{\mathrm{comp}}\}.
 \label{eq:residual}
\end{equation}
Here, $R_{\mathrm{stat}}$ is the infinity norm of the Lagrangian gradient, measuring violation of the stationarity condition. $R_{\mathrm{eq}}$ is the maximum absolute equality-constraint residual, corresponding to defects in the discretized dynamics. $R_{\mathrm{ineq}}$ is the maximum violation of the inequality constraints, including state and input bounds. $R_{\mathrm{comp}}$ measures violation of complementary slackness: the maximum absolute product of an inequality-constraint multiplier and its corresponding constraint margin. These residuals are evaluated over the full prediction horizon. Convergence requires finite residuals with $R\leq10^{-6}$, maintained for 20 further updates.

Of the \RevBaselineSnapshots{} frozen problems formulated by the original MLP-NMPC, \RevBaselineConverged{} satisfy the convergence criterion, whereas \RevBaselineBudget{} do not satisfy it within 2000 updates. For the converged problems, the absolute difference between the first velocity-command increment after one SQP update and that at verified convergence is $\RevRTIConvergedMeanGap{} \pm \RevRTIConvergedSDGap{}$ m/s. Reaching the residual tolerance requires an average of \RevRTIConvergedMeanUpdates{} SQP updates, excluding the 20 additional verification updates.

% Structured comparison: results/solver_revision/structured_convergence/summary.json.
For comparison, we apply the same frozen-problem evaluation procedure to Structured-NMPC at the same \RevBaselineSnapshots{} sampling points, retaining SQP-RTI with zero LM regularization. The MLP-derived delay-predicted initial state, previous command, reference preview, and other OCP parameters are shared, while each model has its own optimizer memory reconstructed by the same recorded-input replay procedure. Structured-NMPC satisfies the convergence criterion in 153 problems; the remaining three exhaust the update budget. Among its converged problems, the corresponding first-increment difference is $0.00021 \pm 0.00079$ m/s, and convergence requires an average of 3.6 updates, excluding the 20 additional verification updates.

The lower convergence rate, larger first-update discrepancies, and greater update requirements of MLP-NMPC indicate poorer numerical convergence under the shared solver settings. This provides a plausible explanation at the optimization level for part of its poorer tracking performance and motivates examining alternative solver configurations and update budgets.

\subsubsection{Offline Solver Evaluation}
To assess whether alternative solver configurations can improve OCP convergence of MLP-NMPC, we conduct an offline grid search over the update strategy, LM regularization parameter $\lambda$, and update budget. The search compares full-step SQP-RTI with line-search SQP using merit-function backtracking, at $\lambda\in\{0,0.05,0.1,0.135,\allowbreak 0.2,0.3,0.4,0.5,\allowbreak 0.7,1.0\}$ and budgets of one, two, or three updates per control step, yielding 60 configurations. All configurations share the model, objective, constraints, and prediction horizon. A budget of $k$ means $k$ successive single-update RTI calls on the same OCP or one line-search SQP call with at most $k$ updates. Each configuration maintains its own optimizer memory during recorded-input replay, with one horizon shift per valid control step; candidate actions do not enter the recorded command history or physical system.

The main test set contains one completed trial for each of the nine reference conditions: the three constant targets and the 6-s and 8-s sine, triangular, and trapezoidal references. These nine trials provide \RevGridMainSteps{} recorded control steps. A separate difficult set includes two fallback-terminated trials and two completed trials containing states at which offline solver tests revealed convergence difficulties. We construct \RevGridDifficultStates{} frozen OCPs from these records: samples near 5, 10, and 20 s from each completed trial, samples near 0.5 s and the final valid control step from each terminated trial, and two additional samples previously identified as numerically difficult. On the main set, we evaluate each solver configuration using the mean absolute error between its first velocity-command increment and that of a numerically converged local reference solution to the same OCP. Errors at sampled control steps are averaged within each trial and then equally across the nine trials. Converged solutions are accepted using the same finite aggregate-residual threshold, $R\leq10^{-6}$, as in the original-solver diagnostic. We also quantify sustained full-bound oscillation by the maximum length, over the main set, of a sequence of consecutive increments alternating between opposite bounds. On the difficult set, each configuration is continued from its own replay-derived optimizer state while the OCP remains fixed, for at most 2000 SQP updates. We report the number of problems that meet the convergence criterion and the median number of updates required among those that converge. As in the original RTI diagnostic, RTI convergence is confirmed over 20 further updates, which are excluded from the reported update counts.

\begin{figure}[tb]
\centering
\includegraphics[width=0.98\linewidth]{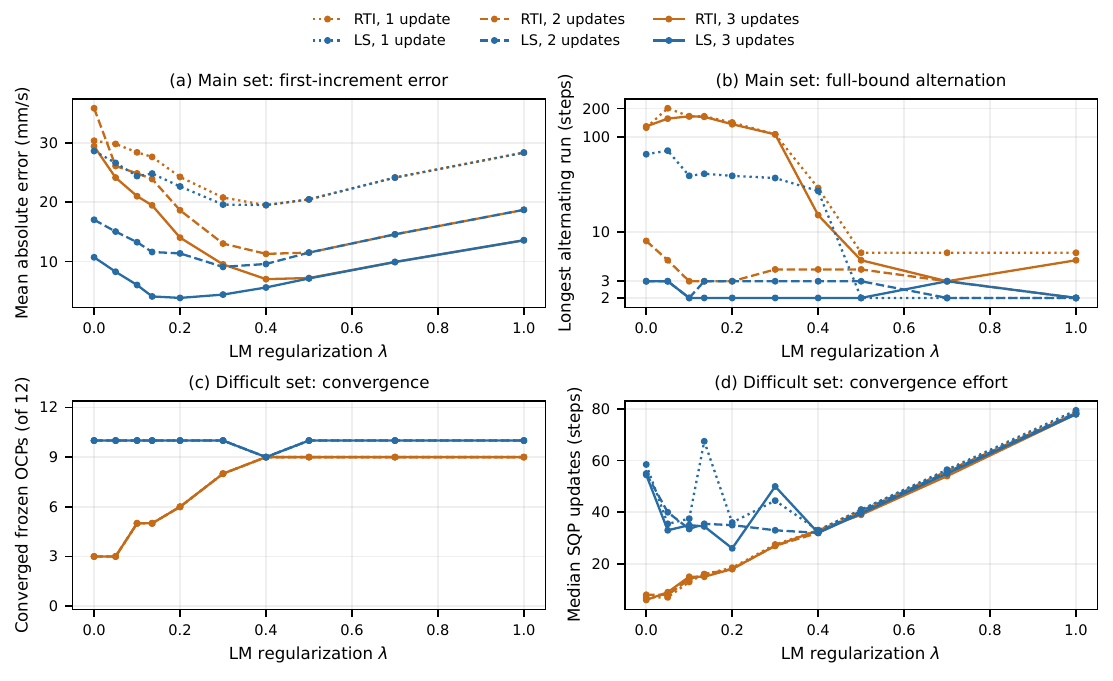}
\caption{Results of the offline grid search. (a) Mean absolute error between each configuration's first velocity-command increment and that of a numerically converged local reference solution to the same OCP in the main set. (b) Maximum number of consecutive velocity-command increments alternating between opposite bounds in the main set. (c) Number of the 12 frozen OCPs satisfying the convergence criterion in the difficult set. (d) Median number of SQP updates required to reach convergence in the difficult set.}
\label{fig:solver-grid}
\end{figure}

The results are reported in Fig.~\ref{fig:solver-grid}. At the same LM regularization parameter $\lambda$ and update budget, line search generally achieves smaller or comparable first-increment errors than RTI (as shown in Fig.~\ref{fig:solver-grid}(a)) and reduces or maintains the maximum length of full-bound alternating sequences (as shown in Fig.~\ref{fig:solver-grid}(b)). It also yields at least as many converged difficult problems, with the clearest advantage at small values of $\lambda$ (see Fig.~\ref{fig:solver-grid}(c)). At these small values of $\lambda$, however, its median update count among converged problems is higher (see Fig.~\ref{fig:solver-grid}(d)). RTI's lower median in this regime concerns only a small subset of successfully solved problems; its action errors remain relatively large, and sustained alternation is prominent with one or three updates per control step. Increasing $\lambda$ to a moderate range reduces RTI's action errors and suppresses sustained alternation, but also increases the update count required for convergence. As $\lambda$ increases further, the finite-budget action errors rise again while the required update counts continue to grow. These trends highlight the difficulty of balancing finite-budget solution accuracy, convergence, and computational effort under real-time constraints.

\subsubsection{Closed-Loop Evaluation}
\begin{figure}[tb]
\centering
\includegraphics[width=0.98\linewidth]{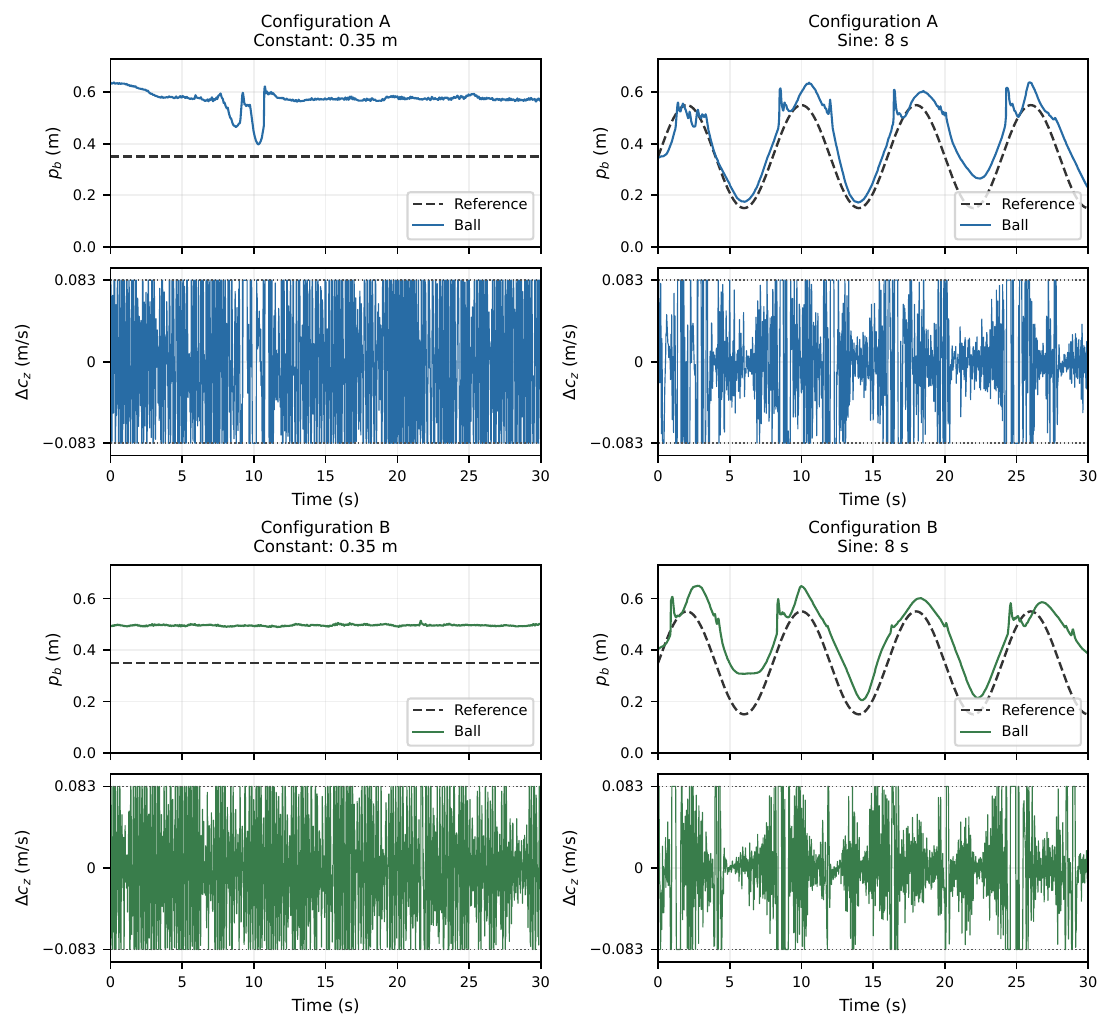}
\caption{Representative completed trials for Configuration A ($\lambda=0.3$, at most two updates) and Configuration B ($\lambda=0.2$, at most three updates).}
\label{fig:solver-online}
\end{figure}

For online evaluation, we consider caps of two and three SQP updates per control step to limit computational demand under real-time constraints. Within each budget, we select the LM regularization parameter $\lambda$ that minimizes the main-set mean first-increment error for line-search SQP:
\begin{itemize}
    \item \textbf{Configuration A:} line-search SQP with $\lambda=0.3$ and at most two updates per control step.
    \item \textbf{Configuration B:} line-search SQP with $\lambda=0.2$ and at most three updates per control step.
\end{itemize}
Each configuration is evaluated on two reference-tracking tasks: a constant target at 0.35~m and a sine reference with an 8-s period, with three trials per task and configuration. The controller changes are confined to the solver settings; the frozen MLP model, objective, constraints, prediction horizon, and delay-compensation settings are retained. The physical apparatus and experimental procedure follow those described in the main paper. The original MLP-NMPC results in the two shared conditions provide the historical baseline.
Tables~\ref{tab:solver-performance} and \ref{tab:solver-runtime} report tracking performance and runtime, respectively.
Fig.~\ref{fig:solver-online} shows the ball trajectories and velocity-command increments from representative completed trials.

\begin{table}[tb]
\centering\small
\caption{Closed-loop tracking performance.}
\label{tab:solver-performance}
% Generated from results/solver_revision/performance.csv
\begin{tabular}{lccc}
\toprule
{\bfseries\boldmath Configuration} & {\bfseries\boldmath SE (cm)} & {\bfseries\boldmath RMSE (cm)} & {\bfseries\boldmath Completed/Attempted} \\
\midrule
\multicolumn{4}{c}{\textbf{Constant: 0.35 m}} \\
\midrule
{\bfseries\boldmath Original MLP-NMPC} & 14.29 $\pm$ 9.82 & - & 6/6 \\
{\bfseries\boldmath Configuration A} & 17.75 $\pm$ 8.56 & - & 3/3 \\
{\bfseries\boldmath Configuration B} & 15.64 $\pm$ 8.03 & - & 3/3 \\
\midrule
\multicolumn{4}{c}{\textbf{Sine: 8 s}} \\
\midrule
{\bfseries\boldmath Original MLP-NMPC} & - & 16.50 $\pm$ 2.90 & 4/6 \\
{\bfseries\boldmath Configuration A} & - & 9.79 $\pm$ 2.10 & 2/3 \\
{\bfseries\boldmath Configuration B} & - & 11.34 ($n=1$) & 1/3 \\
\bottomrule
\end{tabular}

\end{table}

% \begin{table}[tb]
% \centering\small
% \setlength{\tabcolsep}{4pt}
% \caption{Online optimization quality in the twelve line-search attempts. $k$ is the number of SQP updates per control step. Residual and reference statistics use $5\leq t<30$~s. P50/P95 are percentiles of finite aggregate residuals $R$, pooled over control steps. ``NLP converged'' reports the number of steps satisfying $R\leq10^{-6}$ out of the number with finite residuals. ``Regular references'' reports accepted/available regularly sampled frozen OCPs; accepted local references satisfy the same residual tolerance. $E_u$ is the mean absolute difference between the online and reference first velocity-command increments, in m/s, averaged within each trial and then equally across trials. Deadline and final-frame event samples are excluded from the reference counts and $E_u$. ``Run'' is the maximum length, across trials, of consecutive published increments alternating between opposite bounds, using each trial's full observed interval.}
% \label{tab:solver-quality}
% \input{tables/solver_revision/online_quality}
% \end{table}

\begin{table}[tb]
\centering\small
\setlength{\tabcolsep}{4pt}
\caption{Runtime over all observed control steps. P50/P99 are the 50th/99th percentiles of pooled finite per-step times. Command-ready latency spans the start of control-step execution to command readiness, including prediction, optimization, and validation. ``Deadline misses'' counts steps exceeding the 16.67-ms command-ready budget; ``Terminated trials'' counts trials stopped after two consecutive such overruns.}
\label{tab:solver-runtime}
% Generated from results/solver_revision/online_quality.csv
\begin{tabular}{llccccc}
\toprule
{\bfseries\boldmath Configuration} & {\bfseries\boldmath Task} & {\bfseries\boldmath Steps} & {\bfseries\boldmath \shortstack{Solver time (ms)\\P50 / P99}} & {\bfseries\boldmath \shortstack{Command-ready latency (ms)\\P50 / P99}} & {\bfseries\boldmath Deadline misses} & {\bfseries\boldmath Terminated trials} \\
\midrule
{\bfseries\boldmath Configuration A} & {\bfseries\boldmath Constant} & 5403 & 3.90 / 8.67 & 4.47 / 9.90 & 1 & 0 \\
{\bfseries\boldmath Configuration A} & {\bfseries\boldmath Sine} & 4121 & 3.89 / 9.98 & 4.49 / 11.12 & 5 & 1 \\
{\bfseries\boldmath Configuration B} & {\bfseries\boldmath Constant} & 5403 & 5.00 / 10.97 & 5.55 / 12.14 & 1 & 0 \\
{\bfseries\boldmath Configuration B} & {\bfseries\boldmath Sine} & 2905 & 5.33 / 11.84 & 5.89 / 13.03 & 5 & 2 \\
\bottomrule
\end{tabular}

\end{table}

As shown in Table~\ref{tab:solver-performance}, the new solver configurations do not improve the observed mean SE for the constant-reference task: Configuration A yields \RevSEA{}~cm and Configuration B \RevSEB{}~cm, compared with \RevSERTI{}~cm for the original MLP-NMPC. Persistent position offsets remain visible in Fig.~\ref{fig:solver-online}. For the sine reference, however, the mean RMSE among completed trials decreases from \RevRMSERTI{}~cm to \RevRMSEA{}~cm and \RevRMSEB{}~cm, respectively, suggesting a potential benefit for dynamic tracking in the tested condition. Nevertheless, Table~\ref{tab:solver-runtime} shows that computational deadline overruns cause premature termination of sine-tracking trials under both configurations, despite median and 99th-percentile command-ready latencies remaining below the 16.67-ms control-period budget. The observed dynamic-tracking benefit must therefore be considered alongside the computational cost of line search and additional updates.

\subsubsection{Conclusions}
The convergence difficulties identified offline and the limited gains observed in closed-loop tests together support two conclusions about the solver hypothesis:
\begin{itemize}
    \item Under the same solver settings, MLP-NMPC exhibits poorer numerical convergence than Structured-NMPC at the same sampled inputs, as reflected in its lower convergence rate and greater update requirements.
    \item Alternative solver configurations improve offline solution quality and reduce tracking errors in completed sine trials. However, persistent constant-reference errors and deadline-related terminations limit these benefits, making consistent online performance gains difficult under the real-time computational constraints.
\end{itemize}

\subsection{Static-Equilibrium Consistency}
\label{subsec:equilibrium}
% Constant-reference tracking requires both reaching the target and maintaining the conditions that hold the ball there. An equilibrium admitted by the prediction model need not match the physical balance, and an error in that balance can affect the motion anticipated by NMPC. We therefore examine target feasibility, consistency with observed quasi-static states, and the relation of any discrepancy to persistent position offsets. This complements the local-response analysis in Section~V-E of the main paper by examining the predicted balance conditions rather than their local derivatives.

\subsubsection{Feasibility of Target Equilibria}
A static model equilibrium is a state that remains constant under a constant velocity command. At a candidate ball position $p$, it requires
\begin{equation}
 v_b=v_{rz}=\omega=0,\qquad
 \modelacc(p,0,\theta_{\mathrm{eq}},0)=0,\qquad
 c_{z,\mathrm{eq}}=-b/K,\quad \Delta c_z=0,
 \label{eq:equilibrium}
\end{equation}
with the command queue held at $c_{z,\mathrm{eq}}$. Here, $\theta_{\mathrm{eq}}$ is the equilibrium beam angle, $\modelacc$ is the predicted ball acceleration, and $K$ and $b$ are the gain and bias of the identified low-level actuation response. The equilibrium command compensates this bias, so zero drone velocity does not necessarily require a zero velocity command.

Using the frozen Nominal, MLP, and Structured models, we solve the acceleration equation in~\eqref{eq:equilibrium} at 121 positions from 0.05 to 0.65~m in 0.005-m increments, including the three targets. We scan the deployed beam-angle range of $\pm40^{\circ}$, refine bracketed sign changes, and check local minima of the absolute acceleration for roots without a sign change. Accepted roots have an acceleration residual below $10^{-10}$~m/s$^2$. Repeating the scan at twice the angular resolution gives the same detected roots.

The curves in Fig.~\ref{fig:equilibrium} show one detected equilibrium angle for each model at every scanned position. The Nominal model requires zero beam inclination, whereas the learned models require position-dependent inclinations. All three models admit equilibria at the desired ball positions within the configured state and input bounds, with the same equilibrium command of $-b/K$~m/s and zero command increment. Thus, the desired positions do not lack a feasible model equilibrium.

\subsubsection{Consistency with Observed Quasi-Static States}
To assess whether the predicted equilibrium conditions are consistent with physical observations, we identify intervals in which the ball and beam remain approximately stationary. We start from the 18 Structured-NMPC and 18 MLP-NMPC constant-reference trials reported in the main paper, with six trials per controller and target, and divide each record over $5\leq t<30$~s into ten nonoverlapping 2.5-s candidate windows. Both controller datasets are screened using the same criteria. Within each window, the root-mean-square ball velocity and drone vertical velocity must not exceed 0.01 and 0.02~m/s, respectively. To limit both fluctuations and slow drift, the differences between the 95th and 5th percentiles of ball position and beam angle must not exceed 0.01~m and $0.5^{\circ}$, and their absolute linear drift rates must not exceed 0.002~m/s and $0.1^{\circ}$/s, respectively. We also exclude windows with nonfinite states, inadequate sampling coverage, timestamp irregularities, or flagged observation anomalies. Selection therefore uses observed motion and data quality, independently of model predictions, and does not require constant commands. These criteria define observed quasi-static intervals rather than independently verified physical equilibria. The resulting set contains \EqRevStructuredWindows{} windows from \EqRevStructuredTrials{} Structured-NMPC trials and \EqRevMLPWindows{} windows from \EqRevMLPTrials{} MLP-NMPC trials.

\begin{figure}[tb]
\centering
\includegraphics[width=0.88\linewidth]{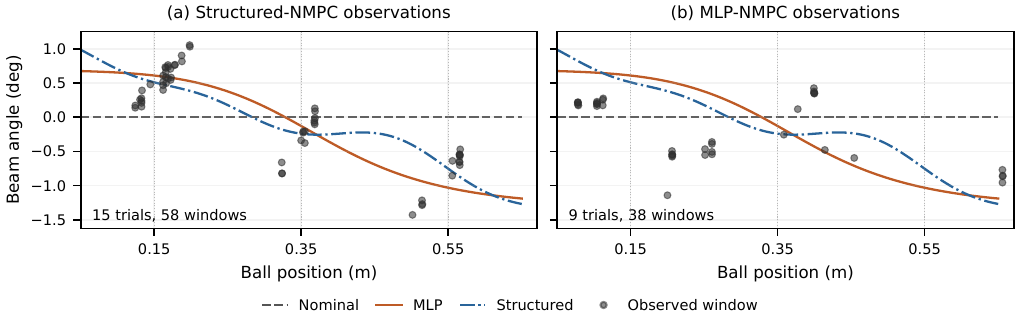}
\caption{Model equilibrium curves and quasi-static observations from (a) Structured-NMPC and (b) MLP-NMPC trials. Both panels use the same three model curves as references. Gray circles mark the median ball position and beam angle of each accepted window. Curves cover 0.05--0.65~m without extrapolation, while all accepted observations are shown. The points are closed-loop quasi-static observations, not independently measured physical equilibria; windows are not independent experimental repetitions.}
\label{fig:equilibrium}
\end{figure}

Fig.~\ref{fig:equilibrium} overlays the selected windows' observed position--angle pairs on the model curves. All three frozen models are evaluated at the same observed states within each window. For a window $W$, we define the equilibrium-angle discrepancy and acceleration-balance residual as
\begin{equation}
\begin{aligned}
 \Delta\theta_W^M
 &=\theta_{\mathrm{eq}}^M(\widetilde p_W)-\widetilde\theta_W,\\
 \varepsilon_W^M
 &=\frac{\displaystyle\int_{t_0}^{t_1}a_b^M(x_{\mathrm{obs}}(t))\,dt
       -[v_b(t_1)-v_b(t_0)]}{t_1-t_0}.
\end{aligned}
\label{eq:equilibrium-consistency}
\end{equation}
Here, $M$ identifies the model, tildes denote window medians, and $t_0,t_1$ are the first and last included observation times. The observed physical state is $x_{\mathrm{obs}}=(p_b,v_b,\theta,v_{rz})^{\top}$. The integral uses the actual sample times and retains the measured velocities. It is an integral consistency check along the observed trajectory, not a free rollout or an assumption of exactly zero physical acceleration.

\begin{table}[tb]
\centering\small
\caption{Consistency with observed quasi-static states.}
\label{tab:equilibrium-consistency}
\begin{tabular}{lcc}
\toprule
\textbf{Model} & \textbf{Angle difference (deg)} & \textbf{Acceleration residual (m/s$^2$)} \\
\midrule
\multicolumn{3}{c}{\textbf{Structured-NMPC observations: \EqRevStructuredTrials{} trials, \EqRevStructuredWindows{} windows}} \\
\midrule
\textbf{Nominal} & $\EqRevStructuredNominalAngleMean{} \pm \EqRevStructuredNominalAngleSD{}$ & $\EqRevStructuredNominalAccelMean{} \pm \EqRevStructuredNominalAccelSD{}$ \\
\textbf{MLP} & $\EqRevStructuredMLPAngleMean{} \pm \EqRevStructuredMLPAngleSD{}$ & $\EqRevStructuredMLPAccelMean{} \pm \EqRevStructuredMLPAccelSD{}$ \\
\textbf{Structured} & $\EqRevStructuredStructuredAngleMean{} \pm \EqRevStructuredStructuredAngleSD{}$ & $\EqRevStructuredStructuredAccelMean{} \pm \EqRevStructuredStructuredAccelSD{}$ \\
\midrule
\multicolumn{3}{c}{\textbf{MLP-NMPC observations: \EqRevMLPTrials{} trials, \EqRevMLPWindows{} windows}} \\
\midrule
\textbf{Nominal} & $\EqRevMLPNominalAngleMean{} \pm \EqRevMLPNominalAngleSD{}$ & $\EqRevMLPNominalAccelMean{} \pm \EqRevMLPNominalAccelSD{}$ \\
\textbf{MLP} & $\EqRevMLPMLPAngleMean{} \pm \EqRevMLPMLPAngleSD{}$ & $\EqRevMLPMLPAccelMean{} \pm \EqRevMLPMLPAccelSD{}$ \\
\textbf{Structured} & $\EqRevMLPStructuredAngleMean{} \pm \EqRevMLPStructuredAngleSD{}$ & $\EqRevMLPStructuredAccelMean{} \pm \EqRevMLPStructuredAccelSD{}$ \\
\bottomrule
\end{tabular}

\end{table}

Table~\ref{tab:equilibrium-consistency} shows nonzero discrepancies for all models. On the Structured-NMPC observations, MLP has a smaller mean angle discrepancy and acceleration residual than Structured. On the MLP-NMPC observations, the two learned models have similar mean angle discrepancies, while MLP again has a smaller mean acceleration residual. To assess sensitivity to the quasi-static selection criteria, we repeat the analysis with all six motion thresholds scaled by 0.75 (stricter) or 1.5 (looser). These changes affect the number of trials with qualifying windows and the relative model rankings based on equilibrium-angle discrepancy. Nevertheless, MLP has a smaller mean absolute acceleration-balance residual than Structured under both threshold settings and for both sources of observations. The two metrics are complementary: their relation depends on the model's angle response, so a smaller acceleration residual alone does not establish a more accurate equilibrium. Taken together, these descriptive comparisons do not show consistently poorer near-static consistency for MLP.

\subsubsection{Relation to Persistent Tracking Offsets}
Of the accepted MLP windows, \EqRevStallWindows{} from \EqRevStallTrials{} trials have mean absolute position error above 5~cm, with a trial-averaged error of \EqRevStallMLPMAEMean{}~cm. To assess whether the model overestimates acceleration toward the target, we multiply $\varepsilon_W^M$ by the sign of $p_g-\widetilde p_W$. A positive value indicates excess predicted acceleration toward the target. For the MLP model, the fraction of these large-offset windows with a positive direction-aligned acceleration-balance residual, averaged equally across trials, is \EqRevStallMLPTowardPctMean{}\%. The corresponding direction-aligned residual is $\EqRevStallMLPAlignedBiasMean{}\pm\EqRevStallMLPAlignedBiasSD{}$~m/s$^2$. Evaluating the Structured model on exactly the same MLP windows gives a positive-residual fraction of \EqRevStallStructuredTowardPctMean{}\% and a direction-aligned residual of $\EqRevStallStructuredAlignedBiasMean{}\pm\EqRevStallStructuredAlignedBiasSD{}$~m/s$^2$, using the same trial-level aggregation.

Both models therefore frequently overestimate acceleration toward the target on these stalled states. Although MLP has a higher positive-residual fraction, its mean direction-aligned residual is not larger. The observed directional bias is compatible with a possible contribution to persistent MLP offsets, but is not specific to MLP. These descriptive comparisons do not establish whether the bias causes insufficient corrective action or explains the controllers' difference in closed-loop performance.

\subsubsection{Conclusions}
The target-feasibility and shared-state consistency checks together support two conclusions about the equilibrium hypothesis:
\begin{itemize}
    \item All three prediction models admit feasible static equilibria at the desired positions. The observed offsets therefore cannot be attributed to those positions being absent from the model's feasible equilibrium set.
    \item Near-static prediction discrepancies accompany persistent MLP tracking offsets, but they are not consistently larger for MLP than for Structured on the same observations. The available evidence does not identify equilibrium mismatch as a distinguishing explanation for MLP's poorer tracking.
\end{itemize}

\begin{figure}[tb]
\centering
\includegraphics[width=0.98\linewidth]{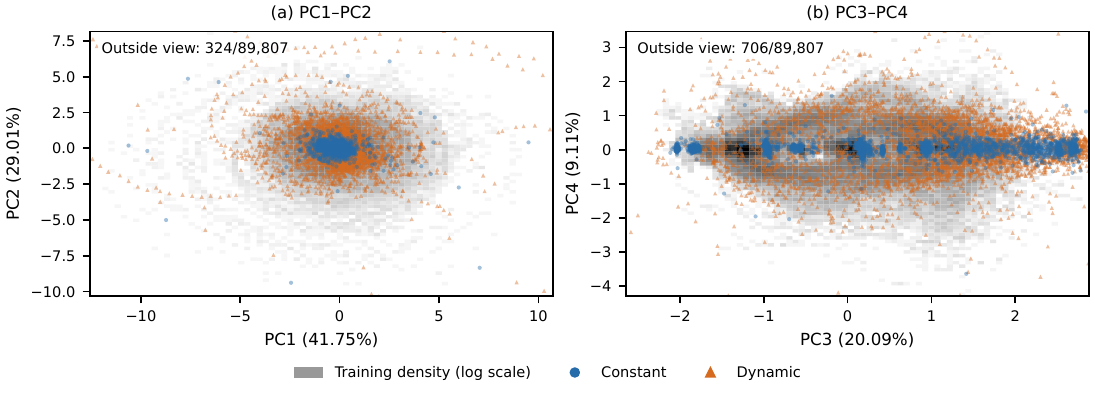}
\caption{Training and closed-loop inputs projected onto (a) PC1--PC2 and (b) PC3--PC4. Axis percentages indicate the training variance captured by each component. The gray background shows training density on a logarithmic scale; blue circles and orange triangles denote constant and dynamic tasks.}
\label{fig:pca}
\end{figure}

\subsection{Input Distribution Shift}
\label{subsec:distribution}
\subsubsection{Data and Evaluation Metrics}
We examine two aspects of the training-deployment input distribution: whether closed-loop inputs are close to recorded training samples, and whether the frequencies of different input values change. The comparison uses the original offline split of \DistRevTrainingTrials{} training, \DistRevValidationTrials{} validation, and \DistRevTestTrials{} test trials, together with the \DistRevTrials{} MLP-NMPC attempts reported in the main paper. The training set provides the reference samples, the validation set determines a distance threshold, and the test set provides an independent offline comparison. The closed-loop data include \DistRevCompleted{} completed trials and the available valid records from \DistRevTerminated{} prematurely terminated trials.

To compare the inputs actually encountered by the MLP, we use its five input features and standardize both offline and closed-loop data with the same frozen training statistics:
\begin{equation}
 z=[p_b,v_b,\theta,\omega,\nomacc]^T,\qquad
 \widetilde z=(z-\mu^{\mathrm{train}})/\sigma^{\mathrm{train}}.
 \label{eq:features}
\end{equation}
Here, $\mu^{\mathrm{train}}$ and $\sigma^{\mathrm{train}}$ are the training mean and standard deviation. This places features with different units on a common scale without adapting the normalization to the closed-loop data. Beam angular velocity $\omega$ and nominal ball acceleration $\nomacc$ are reconstructed from observed physical states using the same equations as in training. All finite closed-loop observations are retained without additional smoothing.

We first measure training-support coverage. For an input $z$, its distance to the nearest valid training sample is
\begin{equation}
 d(z)=\min_{z_i\in\mathcal T}\|\widetilde z-\widetilde z_i\|_2,
 \label{eq:distance}
\end{equation}
where $\mathcal T$ is the set of valid training samples. A smaller distance indicates that a similar combination of all five input features occurs in the training data. This comparison uses the full standardized input space, rather than checking each feature's range separately.
To identify inputs farther from training samples than most held-out offline observations, we set a threshold at the 99th percentile of validation-set distances. We then measure the fraction of observations exceeding this threshold within each trial:
\begin{equation}
 \tau_5=Q_{0.99}\{d(z):z\in\mathcal V\},
 \qquad
 q_5=\frac{1}{n}\sum_{i=1}^{n}\mathbf{1}[d(z_i)>\tau_5].
 \label{eq:coverage}
\end{equation}
Here, $\mathcal V$ contains the valid validation samples, $Q_{0.99}$ denotes the 99th percentile, $n$ is the number of valid observations in a trial, and $\mathbf{1}[\cdot]$ equals one when its condition holds and zero otherwise. The resulting threshold is $\tau_5=\CoverageThreshold$; neither test nor deployment data are used to select it. A low $q_5$ indicates proximity to training samples under this criterion, but does not guarantee accurate predictions.

We next assess changes in input frequency. Even when all observations are close to training samples, a closed-loop trial may spend much more time in particular regions than the training data do. For standardized feature $j$ in trial $r$, we compute the first Wasserstein distance to the training data as
\begin{equation}
 W_{1,j}^{(r)}=
 \int_{-\infty}^{\infty}
 \left|F_{r,j}(s)-F_{\mathcal T,j}(s)\right|\,\mathrm{d}s,
 \label{eq:wasserstein}
\end{equation}
where $F_{r,j}$ and $F_{\mathcal T,j}$ are the empirical cumulative distribution functions of feature $j$ in trial $r$ and the valid training samples, respectively, and $s$ denotes a standardized feature value. The integral measures the area between these two cumulative distribution functions. Larger values indicate greater differences in that feature's distribution.

\subsubsection{Training-Deployment Distribution Comparison}
To visualize the five-dimensional input distributions, we use principal component analysis (PCA), which constructs orthogonal linear combinations of the standardized features, ordered by the training variance they capture. The projection is fitted to the training data only and then applied unchanged to the closed-loop inputs. Fig.~\ref{fig:pca} combines observations from all \DistRevTrials{} MLP-NMPC attempts and shows two complementary views. PC1--PC2 captures \PcaTwoVariance{}\% of training variance; PC3--PC4 shows additional variation omitted from that view, bringing the cumulative variance captured by the first four components to \PcaFourVariance{}\%.
Both projections show substantial visual overlap between closed-loop observations and the training distribution, but the observations occupy these regions differently. Most constant-reference inputs are concentrated near the center of the PC1--PC2 view and remain close to PC4$=0$ across a wider range of PC3 in the second view. Dynamic-task inputs are more dispersed, extending into sparsely populated training regions and beyond the displayed limits (324/89,807 in PC1--PC2 and 706/89,807 in PC3--PC4).

\begin{table}[tb]
\centering\small
\setlength{\tabcolsep}{5pt}
\caption{Training-support coverage in the five-dimensional input space. Median and P95 distances are means of trial-level quantiles; $q_5$ is the trial mean $\pm$ sample SD.}
\label{tab:distribution-support}
% Generated from unchanged distribution_trial_metrics.csv; equal trial weights.
\begin{tabular}{lrrrr}
\toprule
{\bfseries\boldmath Dataset / task} & {\bfseries\boldmath Trials} & {\bfseries\boldmath Median $d$} & {\bfseries\boldmath P95 $d$} & {\bfseries\boldmath $q_5$ (\%)} \\
\midrule
\multicolumn{5}{c}{\textbf{Offline reference}} \\
\midrule
\textbf{Validation} & 6 & 0.096 & 0.296 & 1.0 $\pm$ 1.5 \\
\textbf{Test} & 6 & 0.092 & 0.380 & 2.6 $\pm$ 5.3 \\
\midrule
\multicolumn{5}{c}{\textbf{Closed-loop}} \\
\midrule
\textbf{Constant} & 18 & 0.220 & 0.395 & 15.0 $\pm$ 31.3 \\
\textbf{Dynamic} & 36 & 0.413 & 1.573 & 18.2 $\pm$ 22.2 \\
\bottomrule
\end{tabular}

\end{table}

\begin{table}[tb]
\centering\small
\setlength{\tabcolsep}{5pt}
\caption{Trial-averaged marginal Wasserstein distances to the training data.}
\label{tab:wasserstein}
% Generated from unchanged distribution_trial_metrics.csv; equal trial weights.
\begin{tabular}{lrrrrr}
\toprule
{\bfseries\boldmath Dataset / task} & {\bfseries\boldmath $p_b$} & {\bfseries\boldmath $v_b$} & {\bfseries\boldmath $\theta$} & {\bfseries\boldmath $\omega$} & {\bfseries\boldmath $a_b^{\mathrm{nom}}$} \\
\midrule
\multicolumn{6}{c}{\textbf{Offline reference}} \\
\midrule
\textbf{Validation} & 0.514 & 0.500 & 0.386 & 0.381 & 0.386 \\
\textbf{Test} & 0.537 & 0.474 & 0.439 & 0.369 & 0.441 \\
\midrule
\multicolumn{6}{c}{\textbf{Closed-loop}} \\
\midrule
\textbf{Constant} & 1.489 & 0.450 & 0.544 & 0.349 & 0.550 \\
\textbf{Dynamic} & 1.029 & 0.588 & 0.757 & 0.555 & 0.758 \\
\bottomrule
\end{tabular}

\end{table}

Table~\ref{tab:distribution-support} summarizes training-support coverage of all closed-loop attempts by task. The mean above-threshold fraction is \DistRevConstantQ{}\% across the 18 constant-reference trials and \DistRevDynamicQ{}\% across the 36 dynamic-task attempts, compared with \DistRevValidationQ{}\% for validation and \DistRevTestQ{}\% for test trials. Both closed-loop groups therefore depart more from training support on average, with large standard deviations indicating substantial variation among trials. Nevertheless, the trial-weighted cumulative distance distributions in Fig.~\ref{fig:support}(a) show that most closed-loop observations remain within the prescribed neighborhood of training inputs, indicating substantial support overlap in both task groups despite the increased exceedance fractions.

Table~\ref{tab:wasserstein} complements the coverage analysis by comparing the marginal distributions of individual input features. Relative to the offline reference, constant-reference trials retain comparable Wasserstein distances for ball velocity and angular velocity, but show a larger position-distribution distance, consistent with their concentration in particular regions of the beam. Dynamic trials have higher distances across all five features, with more pronounced increases in position, beam angle, and nominal acceleration. Together with Table~\ref{tab:distribution-support}, these results indicate that substantial training-support overlap can coexist with changes in how frequently input values occur.

\subsubsection{Relation to Tracking Errors}
\begin{figure}[tb]
\centering
\includegraphics[width=0.98\linewidth]{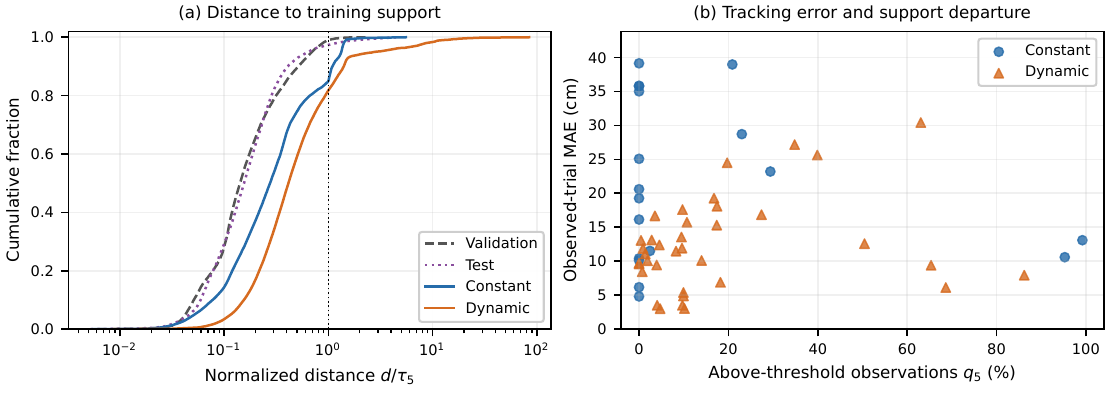}
\caption{Full-space coverage and tracking error. (a) Trial-weighted cumulative distributions of normalized nearest-neighbor distance for the validation, test, constant, and dynamic groups in Table~\ref{tab:distribution-support}; the vertical line marks $d/\tau_5=1$. (b) Observed-trial MAE versus above-threshold fraction for all 54 attempts, with colors and marker shapes distinguishing constant and dynamic tasks.}
\label{fig:support}
\end{figure}

Fig.~\ref{fig:support}(b) compares observed-trial MAE with the above-threshold fraction $q_5$ for constant and dynamic tasks. Each point represents one of the 54 attempts, and both metrics use its available valid observations.
Trials with large tracking errors can still remain well covered by the training inputs, including cases with $q_5=0$. Moreover, the scatter plot does not show a consistent increase in tracking error as $q_5$ increases. These observations indicate that loss of training-support coverage is insufficient as a uniform explanation of poor tracking.

\subsubsection{Conclusions}
Taken together, the coverage, marginal-distribution, and tracking-error analyses suggest two conclusions about the distribution-shift hypothesis:
\begin{itemize}
    \item Closed-loop inputs retain substantial training-support coverage on average, despite differences from the offline reference in exceedance fractions and marginal distributions.
    \item Loss of training-support coverage does not uniformly explain MLP-NMPC's poor tracking, although distribution shift may contribute in individual trials.
\end{itemize}